\documentclass[a4paper]{article}

\usepackage[T1]{fontenc}
\usepackage[utf8x]{inputenc}
\usepackage[english]{babel}
\usepackage{float}
\usepackage{enumerate}

\usepackage[colorlinks=true, allcolors=blue]{hyperref}

\newcommand{\email}[1]{\href{mailto:#1}{\tt{\nolinkurl{#1}}}}
\newcommand{\orcid}[1]{ORCID: \href{https://orcid.org/#1}{\tt{\nolinkurl{#1}}}}

\usepackage[sfdefault,lf]{carlito}
\usepackage[parfill]{parskip}

\usepackage{fancyhdr}
\usepackage{natbib}
\usepackage{authblk}
\usepackage[a4paper,top=3cm,bottom=2cm,left=3cm,right=3cm,marginparwidth=1.75cm]{geometry}

\usepackage{amsmath}
\usepackage{graphicx}
\usepackage{booktabs}
\usepackage{comment}

\usepackage[colorinlistoftodos]{todonotes}

\title{Integrating Multi-Head Convolutional Encoders with Cross-Attention for Improved SPARQL Query Translation}
\author[1]{Yi-Hui Chen}
\author[2,*]{Eric Jui-Lin Lu}
\author[2]{Kwan-Ho Cheng}
\affil[1]{Department of Information Management, Chang Gung University, Taiwan}
\affil[2]{Department of Management Information Systems, National Chung Hsing University, Taiwan}
\affil[*]{Corresponding author: \email{jllu@nchu.edu.tw}} 
\date{2024.8.24}
\begin{document}
\maketitle
\thispagestyle{fancy}

\begin{abstract}

\textbf{{Owner/Author | ACM} {2024}. This is the author's concise and initial version of the work, shared here for personal use only and not for redistribution. The full version has been submitted to ACM Transactions on Information Systems and is currently under review.} \\

The main task of the KGQA system (Knowledge Graph Question Answering) is to convert user input questions into query syntax (such as SPARQL). With the rise of modern popular encoders and decoders like Transformer and ConvS2S, many scholars have shifted the research direction of SPARQL generation to the Neural Machine Translation (NMT) architecture or the generative AI field of Text-to-SPARQL. In NMT-based QA systems, the system treats knowledge base query syntax as a language. It uses NMT-based translation models to translate natural language questions into query syntax. Scholars use popular architectures equipped with cross-attention, such as Transformer, ConvS2S, and BiLSTM, to train translation models for query syntax. To achieve better query results, this paper improved the ConvS2S encoder and added multi-head attention from the Transformer, proposing a Multi-Head Conv encoder (MHC encoder) based on the n-gram language model. The principle is to use convolutional layers to capture local hidden features in the input sequence with different receptive fields, using multi-head attention to calculate dependencies between them. Ultimately, we found that the translation model based on the Multi-Head Conv encoder achieved better performance than other encoders, obtaining 76.52\% and 83.37\% BLEU-1 (BiLingual Evaluation Understudy) on the QALD-9 and LC-QuAD-1.0 datasets, respectively. Additionally, in the end-to-end system experiments on the QALD-9 and LC-QuAD-1.0 datasets, we achieved leading results over other KGQA systems, with Macro F1-measures reaching 52\% and 66\%, respectively. Moreover, the experimental results show that with limited computational resources, if one possesses an excellent encoder-decoder architecture and cross-attention, experts and scholars can achieve outstanding performance equivalent to large pre-trained models using only general embeddings.
\end{abstract}

The question-answering system enables users to obtain answers to their queries. The Knowledge Graph (KG) based question answering system, referred to as KGQA, comprises pre-processing, question type classification, and SPARQL generation modules to handle user inquiries. Initially, pre-processing involves tokenizing the input question. Subsequently, the question type classification module identifies the query forms based on the anticipated answer type. The SPARQL generation module then creates RDF triples relevant to the question. These RDF triples are integrated with the query form to produce the SPARQL syntax. Answers are retrieved using this SPARQL syntax by querying the KG, such as DBpedia. Finally, the answer types are filtered to display the definitive response.

\begin{figure}[H]
    \centering
    \includegraphics[width=10cm]{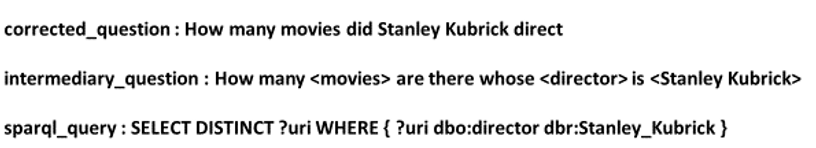}
    \caption{LC-QuAD-1.0 data example}
    \label{fig0-2}
\end{figure}

SPARQL syntax, composed of a query form and RDF triples, serves as the querying language for the datasets. The query form typically includes SELECT DISTINCT, SELECT COUNT, and ASK clauses. An RDF triple consists of a subject, predicate, and object, with the predicate indicating the relationship between the subject and object. For instance, as depicted in Fig. \ref{fig0-2}, corrected\_question signifies the query, while intermediary\_question can be automatically generated based on a preset template incorporating information such as <movies>, <director>, and <Stanley\_Kubrick>. Conversely, sparql\_query denotes the corresponding standard SPARQL for that query on DBpedia. The Query Form for this standard SPARQL is SELECT COUNT, with the RDF triple component being ?uri dbo:director dbr:Stanley\_Kubrick. In this context, ?uri serves as an answer variable in the subject position, dbr:Stanley\_Kubrick is positioned as the object, and dbo:director acts as the predicate, delineating the relationship between the subject and object. 

Based on KGQA, sequence-to-sequence models and attention mechanisms have been proposed for SPARQL generation, exemplified by Transformer\cite{1}, ConvS2S \cite{9}, and significant language models like BERT \cite{2}, T5 \cite{10}, and GPT \cite{11}. The integration of sequence-to-sequence models with attention mechanisms \cite{12, 13} has furthered advancements in generative AI, leading to the development of Text-to-SPARQL methodologies \cite{12, 13, 14, 15, 16, 17}. KGQA research involves converting natural language queries into SPARQL syntax, commonly utilizing techniques such as dependency parsing \cite{20}, entity type tagging \cite{21}, and predefined templates \cite{22, 23}. With the advent of sequence-to-sequence models and attention mechanisms, the efficiency of Neural Machine Translation (NMT) has increased, particularly following the introduction of Transformer and ConvS2S \cite{24}. This has made NMT for Text-to-SPARQL translation increasingly prevalent in recent KGQA system research. The principle of NMT is to facilitate translation between two languages through sequence-to-sequence models, such as translating English to German or English to French. The design concept of NMT in KGQA systems is to treat SPARQL as a language and perform translations between natural language queries and SPARQL using sequence-to-sequence models. 

Yin et al. \cite{13} employed various encoder-decoder architectures such as Transformer and ConvS2S for the conversion of natural language queries into SPARQL, among which ConvS2S achieved the highest BLEU (BiLingual Evaluation Understudy) score of 61.89\% on the LC-QuAD-1.0 dataset. Here, the LC-QuAD dataset and the QALD (Question Answering over Linked Data) dataset \cite{18} are utilized as the primary benchmarks for assessment to evaluate the performance of KGQA systems. Diomedi and Hogan \cite{13} conducted experiments on LC-QuAD-2.0 and WikidataQA using architectures including Transformer, ConvS2S, and LSTM. BLEU is a metric designed to assess machine-translated text automatically. The BLEU score ranges from zero to one and quantifies how closely the machine-translated text aligns with a set of high-quality reference translations. Using ConvS2S as the SPARQL translation model yielded a BLEU score of 65.2\% in the LC-QuAD-2.0 experiments. Although both studies concluded that ConvS2S delivers the best performance in SPARQL translation tasks, Lin and Lu \cite{15} achieved better results using Transformer as the SPARQL translation model, obtaining a 76.32\% BLEU score on the LC-QuAD-1.0 dataset, surpassing Yin et al.'s results by 15\%. Subsequently, Kuo and Lu \cite{16} utilized an LSTM-based encoder-decoder with various cross-attention mechanisms to test performance on datasets such as QALD-9 and LC-QuAD-1.0. Their experiments indicated that their best model surpassed ConvS2S in SPARQL translation tasks.

Inspired by the Transformer \cite{33}, the translation model simplifies the sequence-to-sequence models consisting of the encoder, decoder, and cross-attention modules. To enhance the performances of ConvS2S, the proposed approach incorporates the multi-head attention mechanism from the Transformer to create a Multi-Head Conv encoder. This encoder utilizes filters to extract hidden features from the input sequence, thereby capturing n-gram semantics. As a result, it achieves 83.37\% for BLEU-1 and 46.34\% for exact match scores. An LSTM decoder is employed in the decoder module. In the cross-attention module, three different modules, the Transformer's Multi-Head Cross Attention (MHA), ConvS2S's Multi-Step Cross Attention (MSA), and Multiplicative Cross Attention (MA) \cite{16}, are applied. The experiments reveal that the best performances using MHA in the QALD-9 and LC-QuAD-1.0 datasets are 70.73\% and 77.09\% for BLEU-1, and 34.73\% and 35.74\% for exact match, respectively. 

In terms of the encoder module, within the typical architecture, the Transformer encoder's performance exceeds that of the ConvS2S encoder by approximately 2\% to 4\%. Compared to the typical Transformer decoder module, the LSTM decoder shows a performance improvement of over 2\%.



Two predominant errors, word translation errors and subgraph type errors, are considered when converting natural language queries into SPARQL syntax. An unseen named entity during the inference stage could easily result in translation errors of the named entity. To resolve the word translation errors during the named entity identification stage, NER is utilized to represent named entities. Take the question, "How many movies did Stanley Kubrick direct?" for example; through the NLTK NER Tagger's processing, we generate a modified question: "How many movies did NER direct?" Utilizing NER to replace named entities not only addresses translation errors but also enhances translation accuracy. As for subgraph type errors, NQT (Neural Query Template) serves as the target sequence in the codec architecture, functioning not only as training data during the training phase but also as the target output during the inference phase. Given that the NQTs output by the translation model may contain errors hindering the formation of reasonable subgraphs, we preprocess the standard NQTs in the dataset, derived from the standard SPARQL syntax, into various subgraph patterns. Subsequently, we aim to identify the subgraph types (NQTs) that interrogative sentences should adhere to during the inference phase. We employ this information to review and rectify the NQTs accordingly. The experimental results indicate that these proposed correction mechanisms significantly enhance translation performance in QALD-9 and LC-QuAD-1.0, yielding an approximate 8\% to 12\% increase in both BLEU-1 and Exact match metrics. Notably, the optimal model (Multi-Head
Conv encoder (MHC encoder) combined with MHA), when augmented with the correction mechanism, achieves a substantial leap in translation quality for LC-QuAD-1.0, with BLEU-1 scores reaching 91.61\% \cite{13, 15, 26}, markedly outperforming prior related research.

The end-to-end system employing the optimal translation model (MHC encoder and LSTM decoder combined with MHA) achieved a Marco F1 measure of 52\% and 66\% in the end-to-end system experiments of QALD-9 and LC-QuAD-1.0 respectively, significantly surpassing previous studies \cite{13, 16, 17, 27}. Notably, some of these studies even utilized large pre-trained models like BART. In contrast, our system achieved remarkable performance with a custom-designed encoder-decoder architecture and relatively simple fixed embeddings. This is particularly beneficial for research teams with limited computational resources. Moreover, it allows researchers with more substantial computational resources to build upon and create large pre-trained models for even better performance.

\section{Proposed Method}

\begin{figure}[H]
    \centering
    \includegraphics[width=14cm]{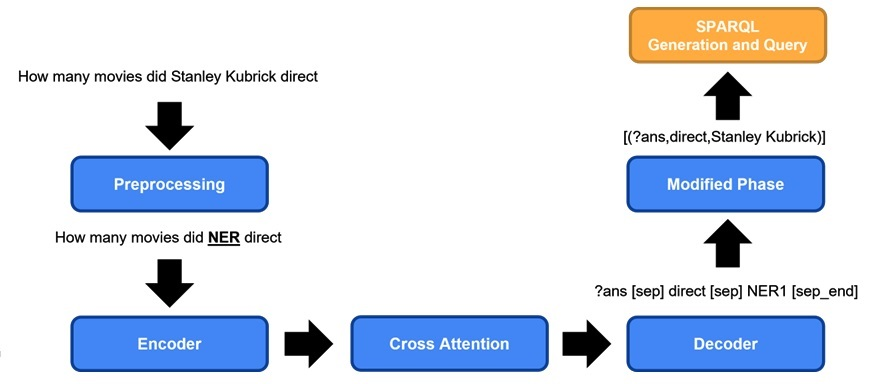}
    \caption{The process of translating NQT using NMT.}
    \label{fig3-2}
\end{figure}

The paper utilizes an NMT model to translate questions into NQT. The NMT model consists of pre-processing, NQT, pre-trained embedding, translate mode, and SPARQL query generation, as shown in Fig. \ref{fig3-2}, and the details are described in Sections \ref{pre-processing}, \ref{NQT}, \ref{pre-trainedembedding} and \ref{translationmodel}, respectively. In the preprocessing stage, named entities are identified in the input question using the NLTK NER Tagger and tagged as NER. In the question "How many movies did Stanley Kubrick direct?", "Stanley Kubrick" is recognized as a named entity and converted into NER, resulting in the input sequence for the translation model: "How many movies did NER direct." 


The input sequence then passes through the three modules of the translation model to obtain the target sequence NQT as one or more sets of RDF triples. The input sequence can be transformed into "?ans [sep] direct [sep] NER1 [sep\_end]". Afterward, the NQT correction mechanism is used to rectify any potential errors in NQT. Finally, the corrected NQT is processed through the SPARQL Generation and Query to generate SPARQL and query the answer to the input question. 

\subsection{Pre-processing} \label{pre-processing}

The pre-processing step encompasses tokenization, NER, and conversion. During the tokenization stage, input sentences are split into tokens using pre-trained embeddings, with whitespaces serving as token separators. During the NER stage, the NLTK NER tagger is employed to identify the named entities in the input questions. For example, the question "How many movies did Stanley Kubrick direct?" is processed to become "How many movies did NER direct?" Also, "How many movies did John Ford direct?" is transformed into "How many movies did NER direct?" After transformation, similar questions can be mapped to the same tagger results, thereby reducing the complexity of the model's learning process.



\subsection{Neural Query Template} \label{NQT}

\subsubsection{Training Stage}\label{Training}

In the previous example, this approach uses two new separator symbols, [sep] and [sep\_end], to replace the original question's commas and periods. In the example question "List the uni. having affiliation with Graham Holding Company and have a campus in Iowa," the target sequence: 

"?ans, affiliation, NER1. ?ans, campus, NER2. ?ans, rdf:type, uni." is transformed into:

ans [sep] affiliation [sep] NER1 [sep\_end] ans [sep] campus [sep] NER2 [sep\_end] ans [sep] rdf:type [sep] uni. [sep\_end]


\subsubsection{Inference Stage} \label{Inferencestage}

The separators are removed in $NQT$, and the words are grouped into three sets to form an RDF triple. Each RDF triple can simultaneously be regarded as an RDF graph consisting of two nodes and one arc (edge). To avoid translation model errors in NQT output, a strategy is employed where standard NQTs in the dataset are pre-simplified into subgraph patterns. During the inference stage, an attempt is made to identify the appropriate subgraph type (represented as $\overline{NQT}$) that the question should conform to. This identified type is then used to correct the $NQT$ accordingly.

\begin{figure}[H]
    \centering
    \includegraphics[width=10cm]{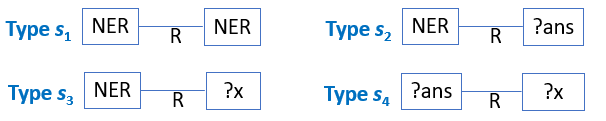}
    \caption{The four basic subgraph types.}
    \label{fig3-4}
\end{figure}

The four basic subgraph types, $s_1$, $s_2$, $s_3$, and $s_4$, have been simplified as shown in Fig. \ref{fig3-4}. In these subgraphs, each node must be either a named entity (NER), an intermediate variable ?x, or the answer variable ?ans, while the edges connecting the nodes must represent property entities and are simplified as R (for relationships). Since the basic subgraph types lack directionality, $s_2$ can represent either (NER, R, ?ans) or (?ans, R, NER).

\begin{figure}[H]
    \centering
    \includegraphics[width=12cm]{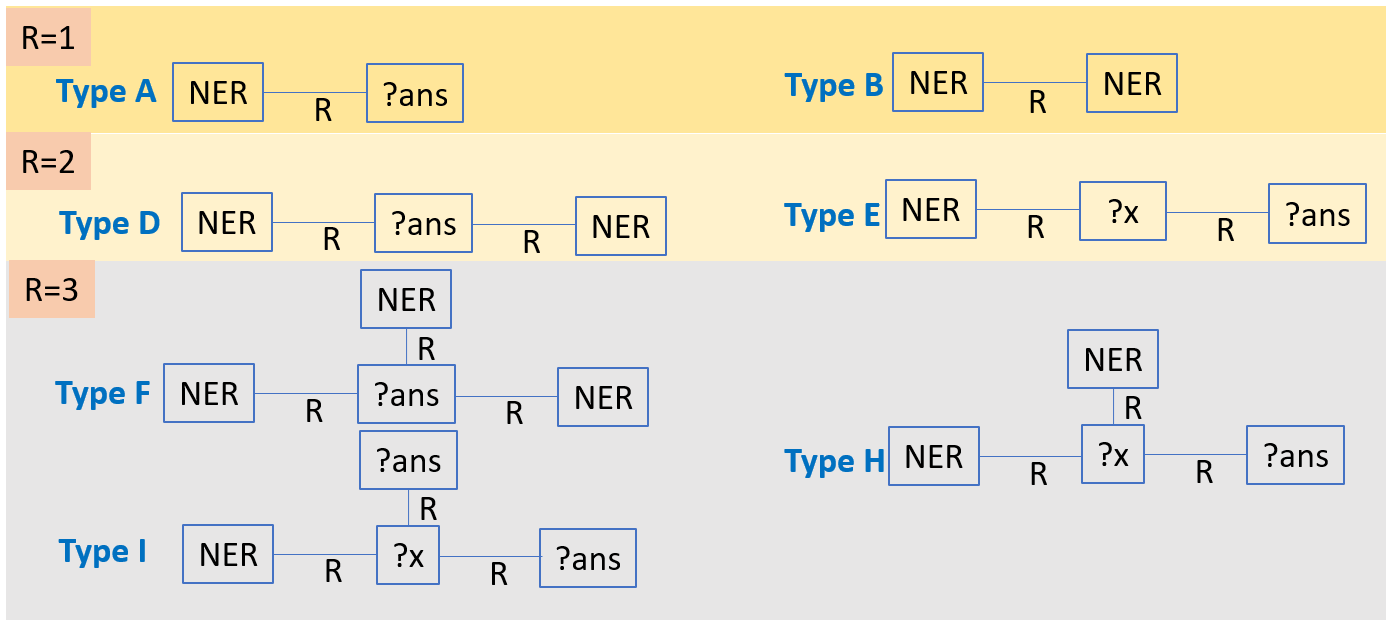}
    \caption{Seven subgraph types.}
    \label{fig3-5}
\end{figure}

Each standard NQT consists of one or more basic subgraph types. Based on the number of property entities, the standard NQTs are derived into seven subgraph types, as illustrated in Fig. \ref{fig3-5}. During the inference stage, the seven subgraph types, each comprising varying quantities of named entities and property entities, can be employed to deduce the subgraph type of $\overline{NQT}$ that corresponds to the question. For example, suppose the question is "Who was in the military unit which played the role of Air interdiction," and its standard NQT is [(?$x$, role, NER1), (?ans, military unit, ?$x$)]. In this case, we can simplify (?$x$, role, NER1) and (?ans, military unit, ?$x$) as $s_3$ and $s_4$ respectively, as described in Fig. \ref{fig3-4}. Since the question has two instances of R and one ?ans, we can combine $s_3$ and $s_4$ into the $E$ type as shown in Fig. \ref{fig3-5}.



To accurately identify the subgraph type in a question, a Multiple Entity Type Tagger (METT) is used to identify question words, named entities, property entities, category entities, and unimportant or stop words represented by $V$, $E$, $R$, $C$, and $N$, respectively. Subsequently, the count of property entities identified by METT is utilized to determine potential $\overline{NQT}$ candidates. As an illustration, when METT detects a single property entity and identifies one named entity, the corresponding $\overline{NQT}$ for the question is designated as A. If METT identifies two named entities in such a scenario, it corresponds to $\overline{NQT}$ B, and so forth. Once $\overline{NQT}$ is determined, it requires corrections if the NQT violates the following two conditions.

\begin{enumerate}[(1)]
    \item The $\overline{NQT}$ matches the subgraph type as the NQT when they share the same entity count.
    \item All the words in the NQT are consistent with the words that appear in the question.
\end{enumerate}

During the correction stage, the NQT is transformed into a format represented by $NQT^*$, aligning it with $\overline{NQT}$. When generating $NQT^*$, the words NER1 and NER2 in NQT are substituted with NER, and the predicate position is replaced with R. For example, If the original NQT is [(NER1, located, ?x)], the resulting $NQT^*$ would be [(NER, R, ?x)]. $NQT^*$ can be mapped to $s_3$ in Fig. \ref{fig3-4}.

Next, in comparing $\overline{NQT}$ with $NQT^*$, both NQT and $NQT^*$ are modified according to the following principle of making minimal changes: retain as much as possible from $NQT^*$ and make the fewest modifications to $NQT^*$. When the basic subgraph type in $NQT^*$ matches a subgraph type in $\overline{NQT}$, that subgraph type is retained. However, if the basic subgraph type $ps_i$ in $NQT^*$ does not appear in $\overline{NQT}$, a basic subgraph type is randomly selected from $\overline{NQT}$ and denoted as $ps_j$. Subsequently, $ps_i$ is replaced with $ps_j$.

In the previous example, $NQT^*$ consists solely of the subgraph type $s_3$, and $\overline{NQT}$ corresponds to D from Fig. \ref{fig3-5}. This indicates that $\overline{NQT}$ includes two subgraph types of $s_2$. Since the $s_3$ subgraph type in $NQT^*$ cannot align with any of the basic subgraph types in $\overline{NQT}$, the modifier randomly selects a basic subgraph type from $\overline{NQT}$ that is not present in $NQT^*$. Adhering to the principle of making minimal changes, $s_3$ could represent either (NER R ?x) or (?x R NER), while $s_2$ could be (NER R ?ans) or (?ans R NER). Consequently, $s_2$ is chosen to modify $s_3$ into $s_2$, entailing the alteration of ?x to ?ans. Furthermore, if this modification is applied to the $i$-th RDF triple in $NQT^*$, the NQT is concurrently updated by replacing the ?x in the $i$-th RDF triple with ?ans.

Secondly, When the number of basic subgraphs in $\overline{NQT}$ and $NQT^*$ are unequal, two cases are employed to correct NQT as follows.

\begin{enumerate}[(1)]
    
    \item When the number of basic subgraph types in $NQT^*$ is fewer than those in $\overline{NQT}$, the missing basic subgraph types in $NQT^*$ are identified by comparing them with $\overline{NQT}$ and subsequently added to $NQT^*$. As an example, after adjusting the basic subgraph types, if $NQT^*$ contains one $s_2$ but $\overline{NQT}$ contains two $s_2$ types, we would need to include an $s_2$ (NER R ?ans) in $NQT^*$, and simultaneously incorporate a corresponding (NER R ?ans) in $NQT$.
    \item If the number of basic subgraph types in $NQT^*$ exceeds those in $\overline{NQT}$, we conduct a comparison between $NQT^*$ and $\overline{NQT}$ and subsequently eliminate the surplus basic subgraph types from $NQT^*$. Certainly, in the given example, if $NQT^*$ comprises two $s_2$ subgraph types and one $s_3$ subgraph type, while $\overline{NQT}$ contains only two $s_2$ subgraph types, then the $s_3$ subgraph type is eliminated from $NQT^*$. Likewise, if the removed subgraph type in $NQT^*$ corresponds to the $i$-th RDF triple, the $i$-th RDF triple in NQT is also deleted.
\end{enumerate}

Words marked as named entities by METT are stored in the array $E_x$. Assuming the question is 'Name the TV show with the distributor as Broadcast syndication and has theme music composed by Primus,' the values of $E_x$ are [Broadcast syndication, Primus]. Next, fill the words in $E_x$ back into the NQT query according to the following rules: 

\begin{enumerate}[(1)]
    \item  Replace NER1 with the first word in $E_x$.
    \item  Replace NER2 with the second word in $E_x$.
    \item  Randomly replace NER with a word from $E_x$ that has not appeared in NQT.
\end{enumerate}

In this example, the corrected NQT is [(NER1, distributor, ?ans), (NER, R, ?ans), (?ans, rdf:type, movies)], so we can obtain the NQT as [(Broadcast syndication, distributor, ?ans), (Primus, R, ?ans), (?ans, rdf:type, movies)].

In cases where the words in the predicate position of NQT are absent in the queried question, the words of the attribute entity from the prior step are stored in the array $R_x$. In the previous example, the array $R_x$ consists of the words [distributor, composed], and as 'distributor' has already been included in NQT, it is disregarded. The word 'composed,' which is not present in NQT, is substituted with the value of R. As a result, the NQT is modified to [(Broadcast syndication, distributor, ?ans), (Primus, composed, ?ans), (?ans, rdf:type, movies)].

 when the predicate position in NQT is 'rdf:type,' and the object position lacks a corresponding word in the queried question, the words from the class entity in the previous step are stored in the array $C_x$. In the ongoing example, $C_x$ is [TV show]. Following the earlier modifications, the resulting NQT is [(Broadcast syndication, distributor, ?ans), (Primus, composed, ?ans), (?ans, rdf:type, movies)]. With $C_x$ containing [TV show], it subsequently becomes [(Broadcast syndication, distributor, ?ans), (Primus, composed, ?ans), (?ans, rdf:type, TV show)].

\subsection{Pre-trained embeddings} \label{pre-trainedembedding}

Three types of embeddings are employed as inputs to capture various semantic information in interrogative sentences: word embeddings \cite{15}, positional embeddings, and NER segment embeddings.

Word embeddings were trained using Word2Vec on all English articles from Wikipedia's October 2020 edition. The NLTK NER Tagger is also utilized to identify all possible named entities in the articles and replace these original words with NER labels. The goal is to enable the model to understand the meaning of NER in questions through pre-trained word vectors.

Positional embeddings are only used in the Transformer encoder and the ConvS2S encoder because these two encoders require positional embeddings to understand the positional information of words in interrogative sentences.

NER segment embeddings are designed similarly to positional embeddings. They assign numbers from 1 to n in the order of appearance of NER tags, with 0 indicating non-NER words. For example, in the sentence "How many movies did Stanley Kubrick direct," after preprocessing, it would become "How many movies did NER direct," where there is only one NER entity. In this case, we represent that NER entity with 1 and non-NER words with 0. Thus, you would get an array like [0, 0, 0, 0, 1, 0], which is then input into an Embeddings Layer to generate NER segment embeddings.

\subsection{Translation model} \label{translationmodel}

\begin{figure}[H]
    \centering
    \includegraphics[width=14cm]{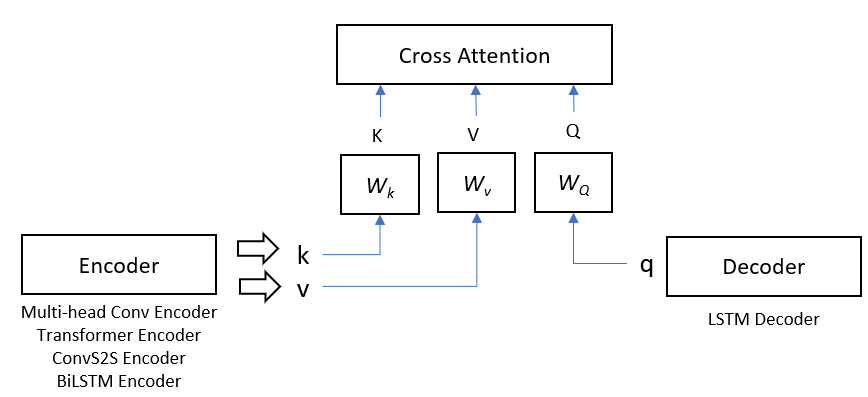}
    \caption{The architecture diagram of the translation model.}
    \label{fig3-6}
\end{figure}

Fig. \ref{fig3-6} illustrates the structure of the translation model, which consists of an encoder, decoder, and cross-attention components. Within the encoder, a novel MHC (Multi-Head Convolution) encoder combines convolutional operations from the ConvS2S encoder with the multi-head attention mechanism found in the Transformer encoder. On the other hand, the decoder utilizes an LSTM decoder. Regarding cross-attention, three distinct types are employed: MHA, MSA, and MA.


To calculate the dot product of the $Q$, $K$, and $V$ matrices, the encoded $k$ and $v$ are first multiplied by weight matrices $W_K$ and $W_V$ respectively, yielding the matrices $K$ and $V$. The decoder-generated $q$ is similarly multiplied by the weight matrix $W_Q$ to obtain $Q$. Subsequently, the dot product of $Q$ and $K$ is computed, and the resulting values pass through a Softmax Layer to determine the attention weights. These attention weights perform a dot product operation with matrix $V$, resulting in the final context information.


\subsubsection{Encoder}

\begin{figure}[H]
    \centering
    \includegraphics[width=10cm]{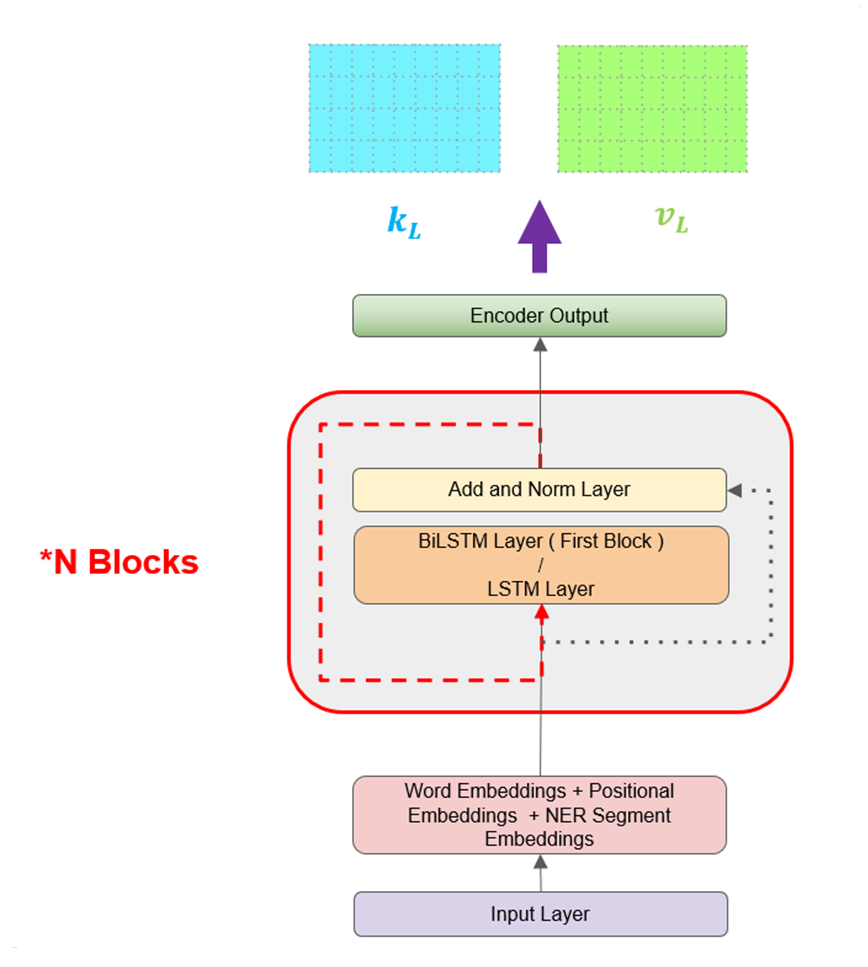}
    \caption{BiLSTM encoder and its output.}
    \label{fig3-7}
\end{figure}

\begin{figure}[H]
    \centering
    \includegraphics[width=9cm]{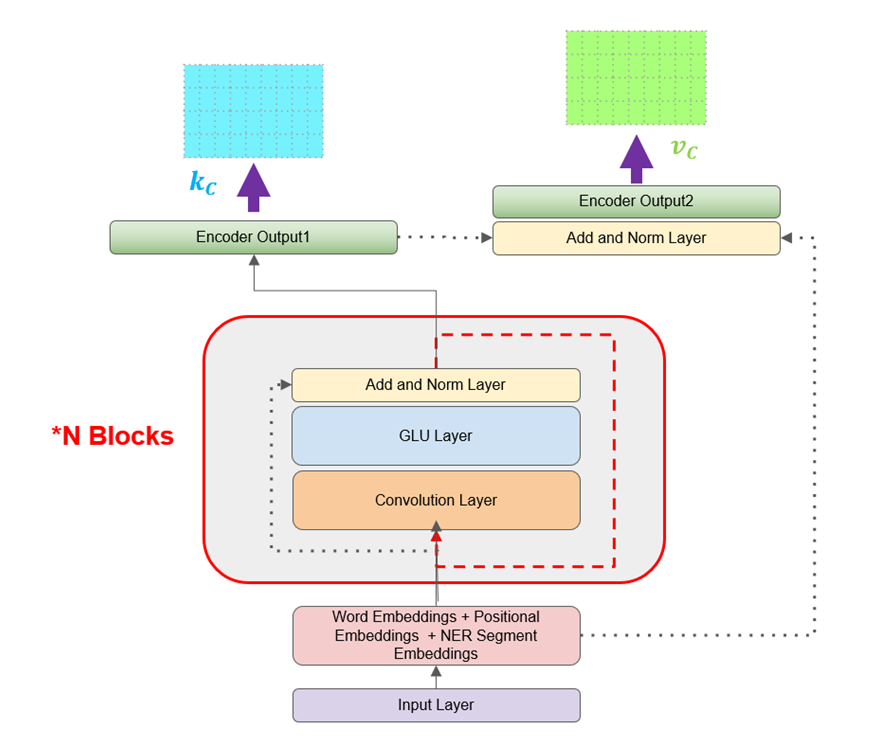}
    \caption{ConvS2S encoder and its output.}
    \label{fig3-8}
\end{figure}

\begin{figure}[H]
    \centering
    \includegraphics[width=9cm]{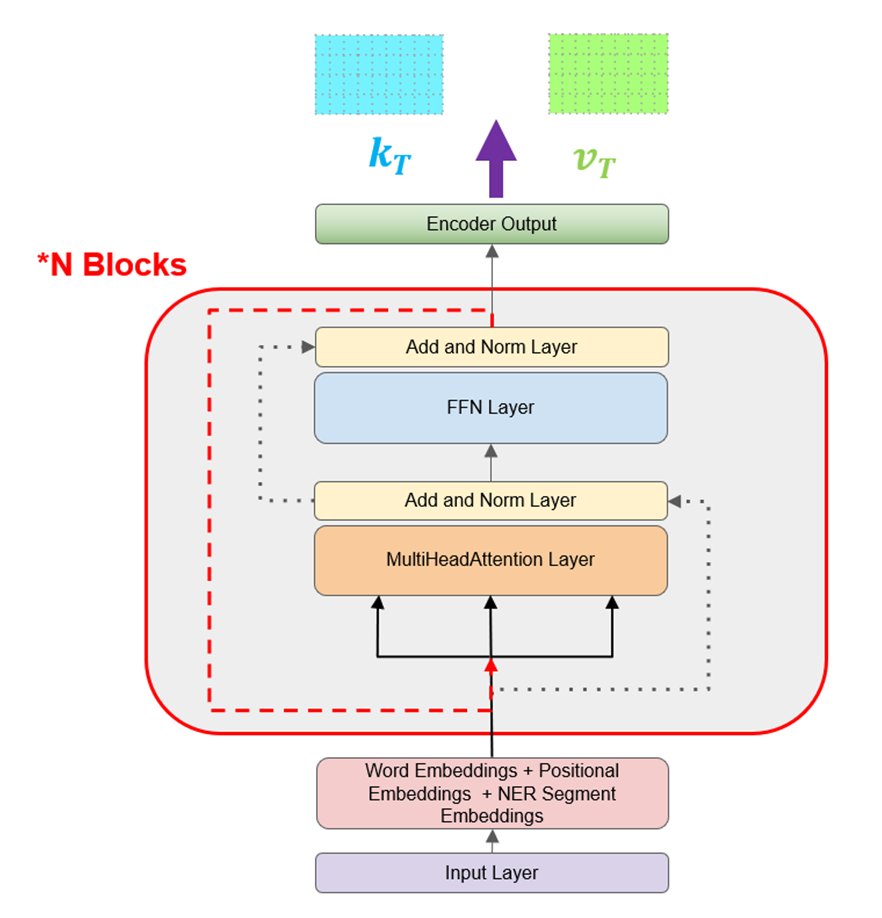}
    \caption{Transformer encoder and its output.}
    \label{fig3-9}
\end{figure}

\begin{figure}[H]
    \centering
    \includegraphics[width=11cm]{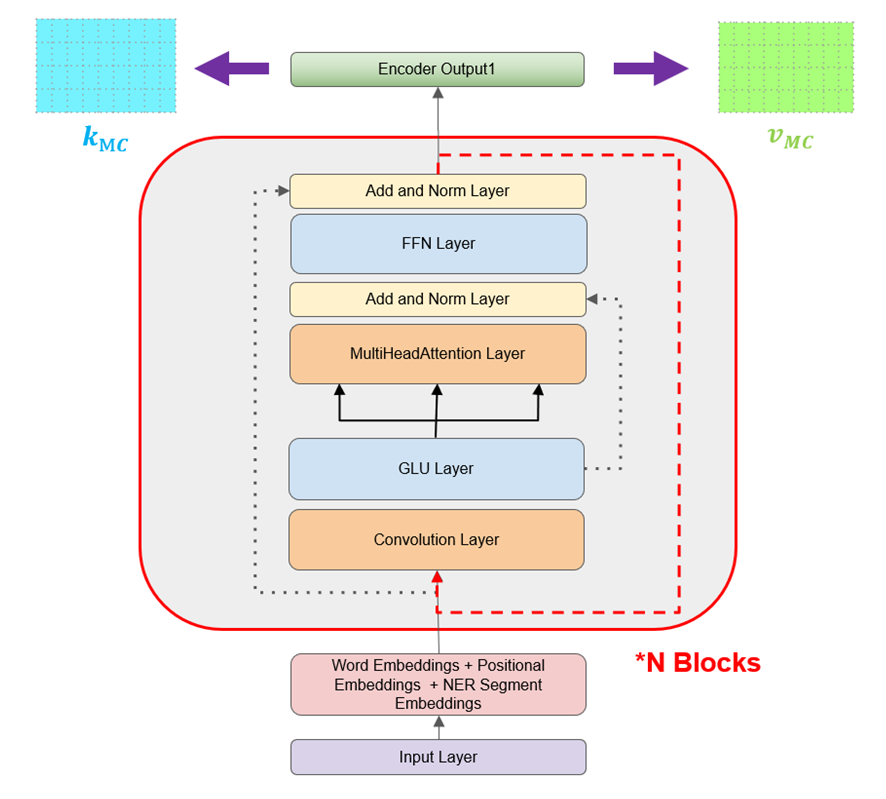}
    \caption{MHC (Multi-Head Convolution) encoder.}
    \label{fig3-10}
\end{figure}

Four encoders are applied to shown in Fig. \ref{fig3-7}, Fig. \ref{fig3-8}, Fig. \ref{fig3-9} and Fig. \ref{fig3-10}. The first encoder is a BiLSTM encoder as shown in Fig. \ref{fig3-7}. The encoder architecture consists of $N$ layers of blocks. The first block is composed of a BiLSTM layer and an add and norm layer, while starting from the second block, it consists of LSTM layers. Since the BiLSTM encoder can capture bidirectional contextual information to enhance the understanding of input sequences and considering the potential loss of original information under multi-layer architecture, we perform addition and normalization (add and norm) on the output of BiLSTM layer and LSTM layer with respect to their inputs, resulting in a high-dimensional representation of an input sequence. Since the output of the last layer of the BiLSTM encoder is only one (i.e., encoder output), it simultaneously serve as the final output \( k_L \) and \( v_L \). These are multiplied by the weight matrices \( W_K \) and \( W_V \) respectively during subsequent cross-attention calculations to generate $K$ and $V$.

Fig. \ref{fig3-8} illustrates the architecture of the ConvS2S encoder, mainly composed of convolution layers, Gated Linear Unit layers (GLU layers), and add and norm layers. The ConvS2S encoder utilizes different filters in N convolution layers to extract local hidden features of n-grams within different receptive fields. Subsequently, these local hidden features pass through a GLU layer, which filters out useful hidden features and removes redundant ones, finally performing addition and normalization with the input of that layer. It is worth noting that the addition and normalization used by the ConvS2S encoder differ slightly from the other two encoders. In practice, it involves scaling the input matrix by a specific value (default $\sqrt{0.5}$). After N layers of computation, the first output, encoder output1, is obtained. Then, encoder output1 is added and normalized with the embeddings layer of the original input to obtain the second output, encoder output2. Considering that in the subsequent cross-attention calculation, the matrix V signifies the matrix that best expresses the original information of the encoder, encoder output2 is used as \( v_C \), and encoder output1 is used as \( k_C \). These are multiplied by the weight matrices \( W_K \) and \( W_V \) respectively during subsequent cross-attention calculations to generate $K$ and $V$.

The architecture of the Transformer encoder is presented in Fig. \ref{fig3-9}, consisting of N blocks composed of multi-head attention layers and feedforward network layers. The outputs of both layers are added and normalized to prevent loss of original information. The main principle of the Transformer encoder is to utilize self-attention operations to compute the dependency relationships between words and generate a high-dimensional representation of the input sequence. After N layers of computation, the output of the last layer of the encoder is used as the encoder output. It simultaneously serves as the final outputs \( k_T \) and \( v_T \), which are multiplied by the weight matrices \( W_K \) and \( W_V \) respectively during subsequent cross-attention calculations to generate \( K \) and \( V \).

The architecture of the MHC (Multi-Head Convolution) encoder is depicted in Fig. \ref{fig3-10}. $n$-gram word embeddings are employed with the MHC to capture more semantic information. Initially, the MHC encoder utilizes convolutional layers to extract the hidden features of n-grams within the receptive field. These features represent the semantics of various n-grams found within the receptive field.

Following this, the MHC encoder leverages the multi-head attention mechanism from the Transformer to calculate attention weights between receptive fields. Additionally, the MHC encoder employs addition and normalization techniques to preserve original information. After undergoing computations across $N$ layers, the output from the final layer serves as the encoder output of the MHC encoder. This output is utilized as both the final values for $K_{MC}$ and $V_{MC}$, which, in subsequent cross-attention calculations, are subjected to multiplication by weight matrices $W_K$ and $W_V$ to generate the matrices $K$ and $V$.

\begin{figure}[H]
    \centering
    \includegraphics[width=12cm]{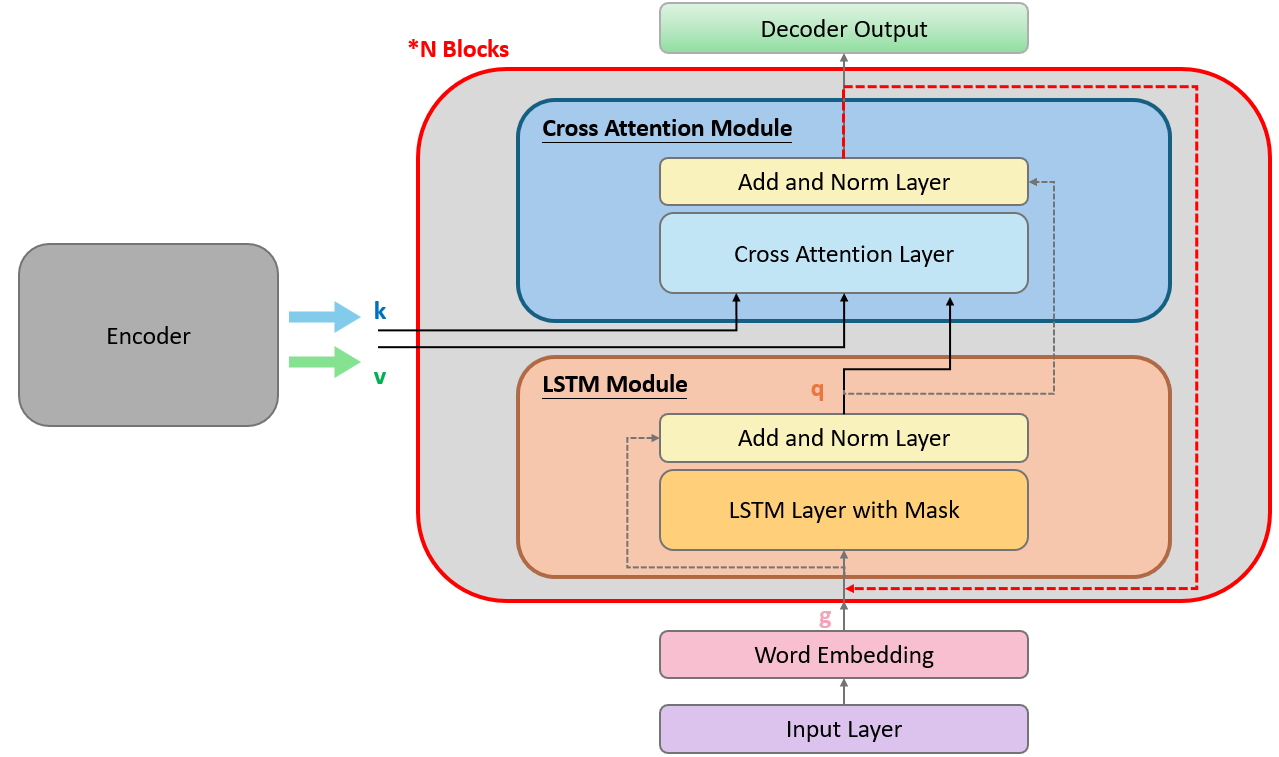}
    \caption{LSTM decoder.}
    \label{fig3-11}
\end{figure}

\subsubsection{Decoder}


The decoder component of our model comprises an $N$-layer LSTM decoder as depicted in Fig. \ref{fig3-11}. Each layer consists of an LSTM module and a Cross Attention module. To capture the contextual information of the target sequence using the LSTM layer, it is then added and normalized with the input of the LSTM module to obtain the output \( q \). Subsequently, in the Cross Attention Module, the encoder outputs \( k \) and \( v \) are inputted along with \( q \) into the Cross Attention Layer to perform cross-attention operations, learning the dependency relationships between the input sequence and the target sequence. Finally, the result is added and normalized with \( q \) to become the output of the \( l \)-th layer of the Cross Attention Module. In Fig. \ref{fig3-11}, \( k \) and \( v \) may have different meanings depending on the encoder used. For example, if paired with a Transformer encoder, \( k \) and \( v \) in Figure 3.11 represent \( k_T \) and \( v_T \) respectively. After N layers of computation, the final output of the decoder, decoder output, is obtained, which is then used to generate the target output for that time step. Additionally, if the Cross Attention Module utilizes Multi-Step Cross Attention (MSA) for computation, it will also use the output of the decoder Embeddings Layer, denoted by the symbol \( g \). The computations of the three types of cross-attention will be explained in Section \ref{CAM}.

\subsubsection{Cross Attention Module} \label{CAM}

\begin{figure}[H]
    \centering
    \includegraphics[width=12cm]{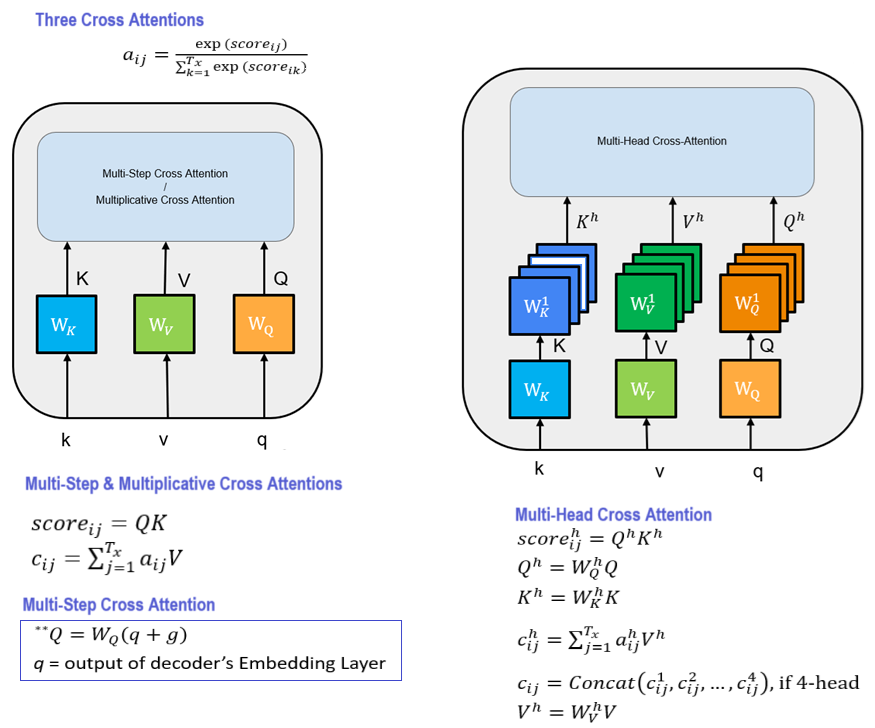}
    \caption{The architecture of three types of cross-attention layers.}
    \label{fig3-12}
\end{figure}

Three distinct cross-attention types are utilized: Transformer's MHA, MSA, and MA as illustrated in Fig. \ref{fig3-12}. The left diagram depicts the structure of MSA and MA, while the right diagram represents the MHA. 

In both diagrams, the primary process involves the multiplication of $q$, $k$, and $v$ by weight matrices $W_Q$, $W_K$, and $W_V$ to generate $Q$, $K$, and $V$ for cross-attention calculations. However, the key difference between the left and right diagrams lies in the subsequent step. In the right diagram, these three matrices are further processed to produce attention head matrices (i.e., $Q^h$, $K^h$, and $V^h$), which are then used for conducting cross-attention calculations.


\begin{figure}[H]
    \centering
    \includegraphics[width=12cm]{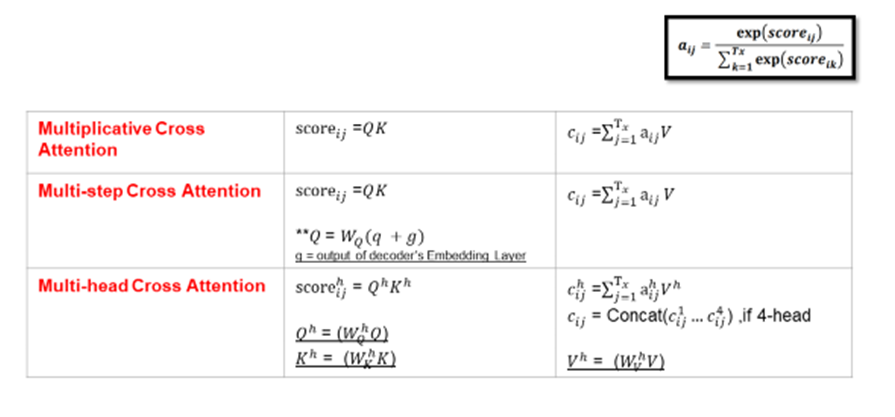}
    \caption{Three types of cross-attention calculation formulas.}
    \label{fig3-13}
\end{figure}

Fig. \ref{fig3-13} elaborates three examples of distinct cross-attention layers. The computation process for MA involves calculating the dot product of $Q$ and $K$ to derive the attention scores for the $l$-th layer, denoted as $score_{ij}$. Here, $i$ corresponds to the $i$-th input from the encoder ($i=1\cdot M$, where $M$ is the length of the input sequence), and $j$ represents the $j$-th output from the decoder ($j=1\cdot T$, where $T$ is the length of the target sequence). Subsequently, the scores $score_{ij}$ are transformed into attention weights for that layer, denoted as $a_{ij}$, by applying the softmax function. Following this, the context information $c_{ij}$ for the $l$-th layer is computed by performing a dot product between $V$ and $a_{ij}$. In Fig. \ref{fig3-13}, $T_x$ represents the current time step, indicating the prediction of the third output word. In this example, $T_x$ is equal to 3. Ultimately, upon completing $N$ layers of interactive attention operations, the decoder produces a word as the output for that particular time step.


The computation method for MSA is similar to MA, with the distinction in calculating attention $score_{ij}$ for the $l$-th layer. It involves deriving $Q$ by summing and normalizing $q$ with the output $g$ from the Embeddings Layer of the decoder. $Q$ is then subjected to multiplication (dot product) with $K$ to obtain the $score_{ij}$. Subsequently, $score_{ij}$ is transformed via the Softmax function to yield the attention weights $a_{ij}$ for that specific layer. Finally, the context information $c_{ij}$ for the $l$-th layer is obtained by conducting a dot product between $V$ and $a_{ij}$.

The design of MHA is notably distinct, and its principle revolves around performing cross-attention operations utilizing multiple attention heads. These attention heads are created by multiplying $Q$, $K$, and $V$ with weight matrices $W_Q^h$, $W_K^h$, and $W_V^h$, yielding $Q^h$, $K^h$, and $V^h$ for each individual attention head. The dimensions of $W_Q^h$, $W_K^h$, and $W_V^h$ are determined by dividing the dimensions of $Q$, $K$, and $V$ by the number of attention heads. For instance, if the dimensions of $Q$, $K$, and $V$ are 256, and there are 4 attention heads (denoted as $h=$1,2,3,4), then $W_Q^h$, $W_K^h$, and $W_V^h$ will transform $Q$, $K$, and $V$ into $Q^h$, $K^h$, and $V^h$ with dimensions of 64 (i.e., 256 divided by 4).

Subsequently, the attention scores $score_{ij}^h$ for each attention head in the $l$-th layer are calculated using the dot product of $Q^h$ and $K^h$. Following this, $score_{ij}^h$ is subjected to the Softmax operation to derive the attention weights $a_{ij}^h$ for that specific attention head. $V^h$ is then multiplied (dot product) by $a_{ij}^h$ to generate the context information $c_{ij}^h$ for that particular attention head. Finally, all the context information $c_{ij}^h$ from the individual attention heads is concatenated to obtain the context information $c_{ij}$ for the $l$-th layer.

\section{Experiments} \label{experiments}

The development work was conducted using Python on a Linux system. Four types of encoders were implemented, including the MHC encoder, Transformer encoder \cite{24}, ConvS2S encoder \cite{35}, and BiLSTM encoder \cite{1, 9}. Additionally, an LSTM decoder with three types of cross-attention mechanisms \cite{1, 16, 25} was developed using TensorFlow 2.11.

This approach demonstrates the use of a comprehensive and diverse set of tools and frameworks in the development of the system. Incorporating multiple encoder types (MHC, Transformer, ConvS2S, BiLSTM) and LSTM decoders with various cross-attention mechanisms indicates a focus on experimenting with different neural network architectures to optimize the system's performance. TensorFlow 2.11, a widely used and powerful open-source library for machine learning, provides a robust platform for implementing these complex models. Additionally, the choice of a Linux environment is suitable for such computational tasks, as it offers advantages in terms of performance and flexibility.

\subsection{Datasets}

The evaluation benchmarks used included the QALD-9 dataset \cite{9} and the LC-QuAD-1.0 dataset \cite{25}. Both datasets provide English questions and corresponding standard SPARQL (Gold SPARQL) queries on DBpedia. QALD-9, proposed by the Question Answering over Linked Data conference, is a widely recognized benchmark for evaluating natural language question-answering systems. LC-QuAD-1.0 is a larger dataset than QALD-9, containing a total of 5000 questions and their corresponding standard SPARQL queries on DBpedia. This dataset is further divided into 4000 questions for training and 1000 questions for testing. Unlike QALD-9, LC-QuAD-1.0 was generated using fixed rules and templates and then fine-tuned through manual adjustments to create the questions. It's worth noting that LC-QuAD-1.0 was designed using the DBpedia version from April 2016, and for consistency, this study utilized the same version of DBpedia for both the QALD-9 and LC-QuAD-1.0 experiments.


Additionally, it was found that some of the standard SPARQL queries provided in these datasets did not correctly retrieve answers from the knowledge base, so these were excluded from the study. Comparative and superlative questions were also excluded: in QALD-9, there were 51 such questions in the training set and 9 in the test set, while in LC-QuAD-1.0, there were 20 in the training set and 9 in the test set. The final number of questions in the training and test datasets can be referenced in Table \ref{table4-1}.



\begin{table}[H]
\caption{Data Counts of QALD-9 and LC-QuAD-1.0}
\begin{tabular}{|c|c|c|} 
  \hline
  & Number of Train dataset & Number of Test dataset \\ 
  \hline
 QALD-9	&302	&82 \\ 
  \hline
  LC-QuAD-1.0	&3967	&991 \\ 
  \hline
\end{tabular}\label{table4-1}
\end{table}

\subsection{Evaluation metrics}

The experiments use the BLEU score \cite{16} and the Exact Match metric to evaluate the similarity between the system's prediction results and the ground truth. The BLEU score, originally designed for machine translation evaluation, is a widely used metric for comparing the similarity between predicted and reference text. It provides a score between 0 and 1, where a higher score indicates better translation quality. The calculation formulas for BLEU-1 \cite{17, 37} are presented in equations (\ref{eq4-1}) and (\ref{eq4-2}):

\begin{equation}
    \label{eq4-1}
    BP=\begin{cases}
    1, & c \leq r \\
    e^{(1-\frac{r}{c}} & c < r \\
    \end{cases}
\end{equation}

\begin{equation}
    \label{eq4-2}
    BLUE - 1 = BP \times exp(log(p_1))
\end{equation}


The Brevity Penalty (BP) accounts for the length difference between the generated prediction result and the reference ground truth. In this formula, \( c \) represents the length of the prediction result, and \( r \) is the length of the ground truth. When the length of the prediction result (\( c \)) is shorter than the length of the ground truth (\( r \)), a penalty weight is calculated using the formula \( e^{(1-r/c)} \). If the length of the prediction result (\( c \)) is greater than or equal to the length of the ground truth (\( r \)), then (\( BP \)) is set to 1. This penalty helps account for cases where the generated response may be shorter than the reference, ensuring a fair translation quality evaluation. Next, we need to calculate the number of matches for all 1-grams (single words) between the prediction result and the ground truth, denoted as \( p_1 \). After applying logarithmic and exponential operations to \( p_1 \), it is multiplied by BP to obtain the BLEU-1 score.

To compute the BLEU-1 score, in this example, the prediction results and the ground truth contain three words, so \( c \) equals \( r \), and thus BP is 1. Next, the number of 1-gram matches (\( p_1 \)) between the prediction result and the ground truth is counted. Since the prediction result matches the ground truth in terms of "?ans" and "NER1" but not "direct," \( p_1 \) would be \( \frac{2}{3} \). Finally, you multiply BP by \( \exp(\log(p_1)) \) to obtain the final BLEU-1 score, which in this case is approximately 0.839. This score reflects the machine translation quality with respect to 1-gram matches between the prediction and the reference.

In the given example, if the standard answer remains unchanged but the prediction result is (NER1, movies, ?ans), the BLEU-1 score remains unaffected despite the difference in the subject and object content. Similarly, even when the prediction result is (NER1, direct, ?ans), the BLEU-1 score remains 1. This demonstrates that BLEU-1 solely counts the number of word matches between the prediction result and the standard answer without considering the order of words. Since different word orders signify distinct RDF triples, a BLEU-1 score of 1 might still result in different queries for answers. Therefore, relying solely on BLEU-1 to determine if the prediction result matches the standard answer may not be sufficient.

A more stringent metric, Exact Match, signifies that the prediction result and the standard answer are identical, and it assigns a value of 1 in such cases. If there is any disparity between them, the metric yields a value of 0. NQT implies that both have matching RDF triples in terms of number and content, indicating that the subject, predicate, and object are all identical.

\subsection{Experimental Results of the Translation Model}


Before NQT correction, the MHC-LSTM model paired with MHA achieved a BLEU-1 score of 83.37\% and an Exact match score of 46.34\% in the LC-QuAD-1.0 experiment. After applying the NQT correction mechanism, these scores improved significantly to 91.61\% for BLEU-1 and 57.09\% for Exact match. Similar improvements were observed in the QALD-9 experiments. Across all our experiments, various translation models demonstrated substantial enhancements in both BLEU-1 and Exact match scores, typically in the range of 8-12\%, after the NQT correction. These results highlight the effectiveness of the NQT correction mechanism in improving translation quality, irrespective of the hybrid model used. Therefore, all subsequent BLEU-1 and Exact match scores analyses are based on results obtained after NQT correction.

\subsubsection{Performances on QALD-9}

The translation model experiments revolve around two key aspects: the model architecture and the cross-attention mechanism. Separate evaluations are conducted to assess the performance of various cross-attention mechanisms while keeping the model architecture constant. Additionally, we have explored how different model architectures perform when using a consistent cross-attention mechanism.

\begin{figure}[H]
    \centering
    \includegraphics[width=10cm]{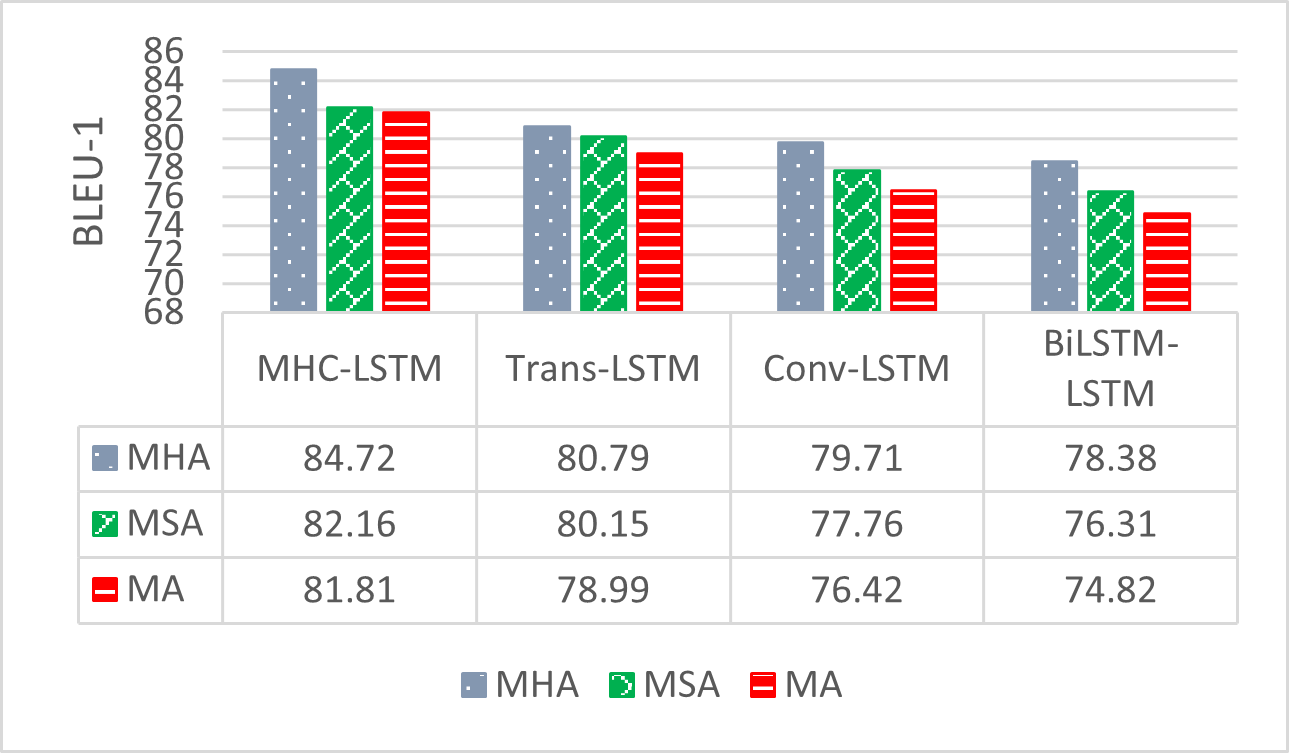}
    \caption{BLEU-1 on QALD-9 with Different Cross-Attention Mechanisms under a Fixed Model Architecture.}
    \label{fig4-1}
\end{figure}

\begin{figure}[H]
    \centering
    \includegraphics[width=10cm]{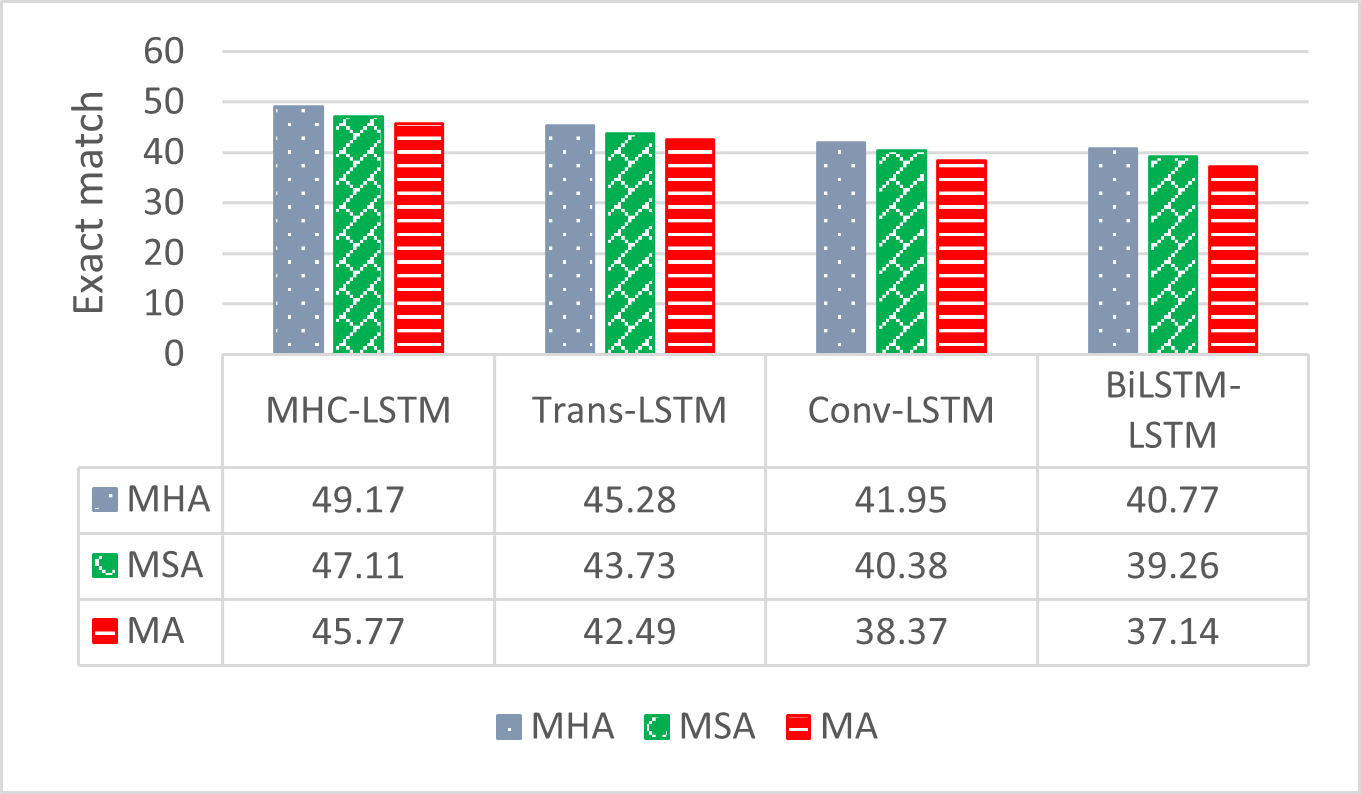}
    \caption{Exact Match on QALD-9 with Different Cross-Attention Mechanisms under a Fixed Model Architecture.}
    \label{fig4-2}
\end{figure}

Figs. \ref{fig4-1} and \ref{fig4-2} illustrate the performance of BLEU-1 and the Exact Match for different cross-attention mechanisms within a fixed model architecture. Specifically, when utilizing the Trans-LSTM model architecture with MHA, we observed BLEU-1 and Exact match scores of 80.79\% and 45.28\%, respectively. MHA outperformed MSA and MA by approximately 1\% and 2\% in BLEU-1 and about 2\% and 3\% in Exact match, respectively.

Similarly, under the fixed Conv-LSTM model architecture, the performance with different cross-attention mechanisms closely mirrored that of the Trans-LSTM model. The combination with MHA achieved the highest BLEU-1 and Exact match scores, reaching 79.71\% and 41.95\%, respectively, surpassing the performance of MSA and MA. In the case of the BiLSTM-LSTM model with various cross-attention mechanisms, the best performance in BLEU-1 and Exact match scores was once again achieved with MHA, reaching 78.38\% and 40.77\%, respectively. MSA followed closely, while MA performed the least effectively. 

The results indicate that all three model architectures performed best when paired with MHA, followed by MSA, and MA performed the least effectively. Among them, the Trans-LSTM model paired with MHA achieved the best performance. Additionally, our proposed MHC-LSTM model, when paired with different cross-attention mechanisms, also performed best with MHA, achieving BLEU-1 and Exact match scores of 84.72\% and 49.17\%, respectively. MSA was the second-best, and MA was the least effective. Comparing MHC-LSTM with other hybrid architectures, MHC-LSTM paired with MHA was found to have the best performance.

\begin{figure}[H]
    \centering
    \includegraphics[width=12cm]{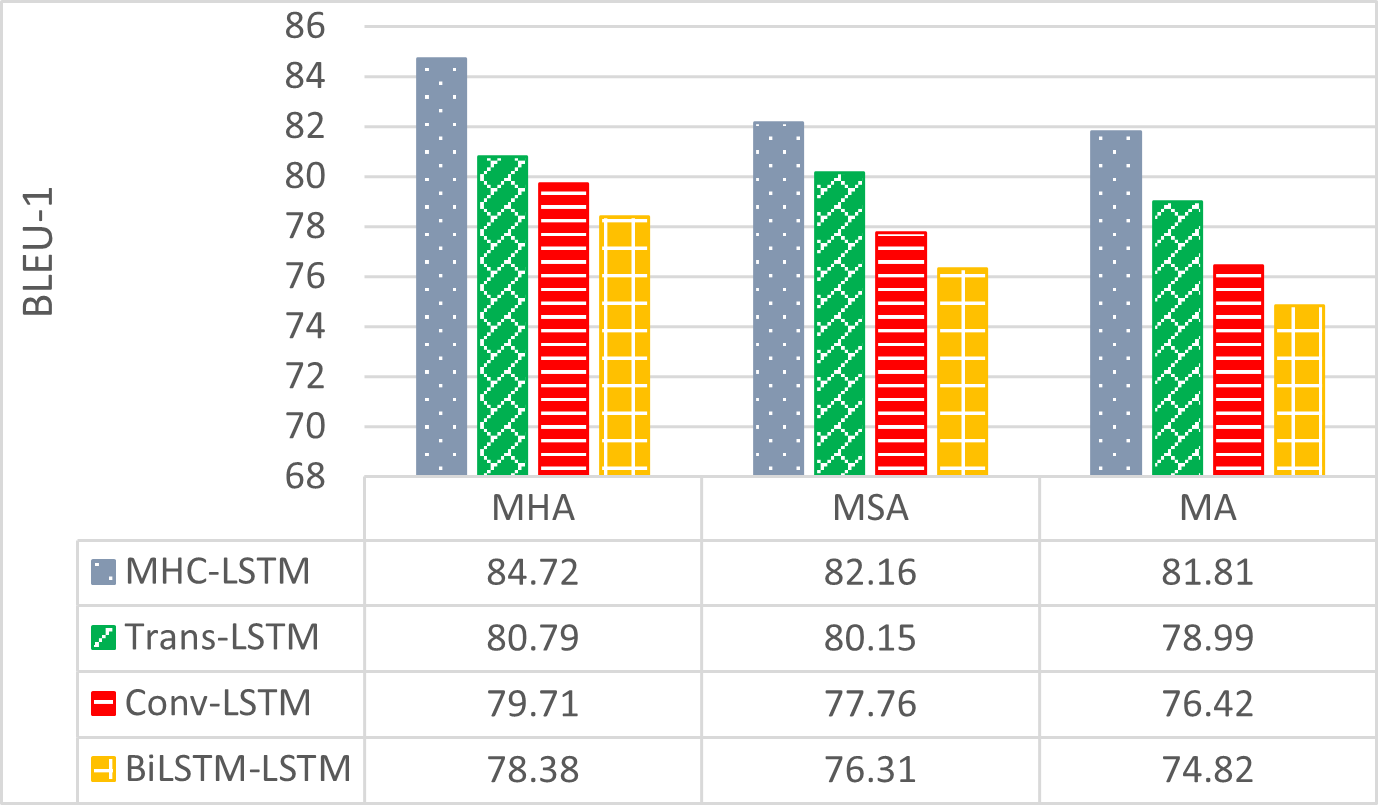}
    \caption{BLEU-1 on QALD-9 with Different Model Architectures under a Fixed Cross-Attention.}
    \label{fig4-3}
\end{figure}

\begin{figure}[H]
    \centering
    \includegraphics[width=12cm]{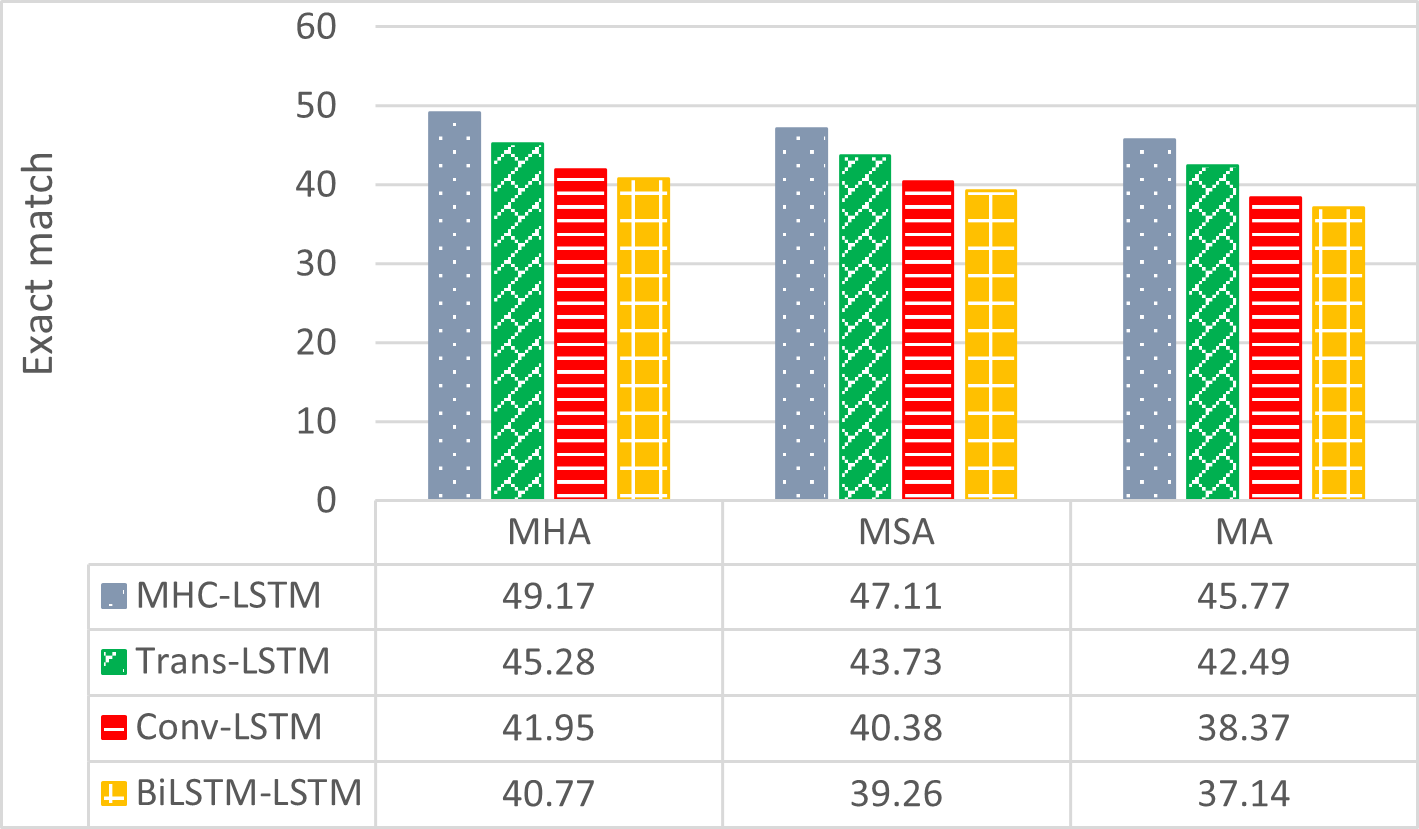}
    \caption{Exact Match on QALD-9 with Different Model Architectures under a Fixed Cross-Attention.}
    \label{fig4-4}
\end{figure}

The experimental results from Figs. \ref{fig4-1} and \ref{fig4-2} were reorganized to compare the performance of different model architectures when paired with a fixed cross-attention mechanism. The results of this analysis are presented in Figs. \ref{fig4-3} and \ref{fig4-4}, and summarized as follows. 

\begin{enumerate}[(1)]
    \item When any single type of Cross Attention is used, the MHC-LSTM model outperformed all other models, followed by Trans-LSTM. Conv-LSTM and BiLSTM-LSTM exhibited lower performance in comparison.
    \item Irrespective of the encoder architecture, MHA consistently outperformed MSA and MA, with MSA showing superior performance compared to MA.
    \item  The figures presented in Figs. \ref{fig4-1} to \ref{fig4-4} demonstrate that when a fixed encoder architecture is utilized, various cross-attention mechanisms can result in a maximum difference of approximately 3-4\% in both BLEU-1 and Exact match scores. Conversely, different encoder architectures can lead to a more substantial variance when employing a fixed cross-attention mechanism, with a maximum difference of approximately 6-7\% in BLEU-1 and 8-9\% in Exact match scores. These findings indicate that the encoder architecture has a more pronounced impact on performance.
\end{enumerate}


In summary, for the QALD-9 experiment, the MHC-LSTM model paired with MHA is the optimal model architecture. Furthermore, MHA is the most effective cross-attention mechanism, regardless of the chosen architecture. This comprehensive assessment highlights the significance of model architecture and cross-attention mechanism selection in achieving peak performance in natural language processing tasks.


As the MHC encoder is an enhancement of the ConvS2S encoder, and the ConvS2S encoder's receptive field size can be adjusted using kernel sizes (e.g., a kernel size of 3 corresponds to a 3-gram receptive field), this study investigated the effects of varying kernel sizes. The outcomes of these experiments are summarized in Table \ref{table4-10}.

In Table \ref{table4-10}, among the various kernel sizes tested, MHC-LSTM with MHA and a kernel size of 3 demonstrated the best translation performance, with BLEU-1 and Exact match scores of 84.72\% and 49.17\%, respectively. This performance outperformed the kernel size of 5 by nearly 1.5\% and was almost 3\% higher than the least effective kernel size of 7.




The performance improvements when comparing the hybrid models to their original counterparts are summarized in Table \ref{table4-10}. The results show that the hybrid models generally outperform the original models. Trans-LSTM with MHA exhibited a slight improvement of 0.63\% in BLEU-1 and approximately a 1\% increase in Exact match compared to Transformer. Among the three models, Conv-LSTM with MHA surpassed ConvS2S, achieving higher scores in both BLEU-1 and Exact match, with an improvement of about 2\%.




\subsubsection{Performances on LC-QuAD-1.0}

\begin{figure}[H]
    \centering
    \includegraphics[width=12cm]{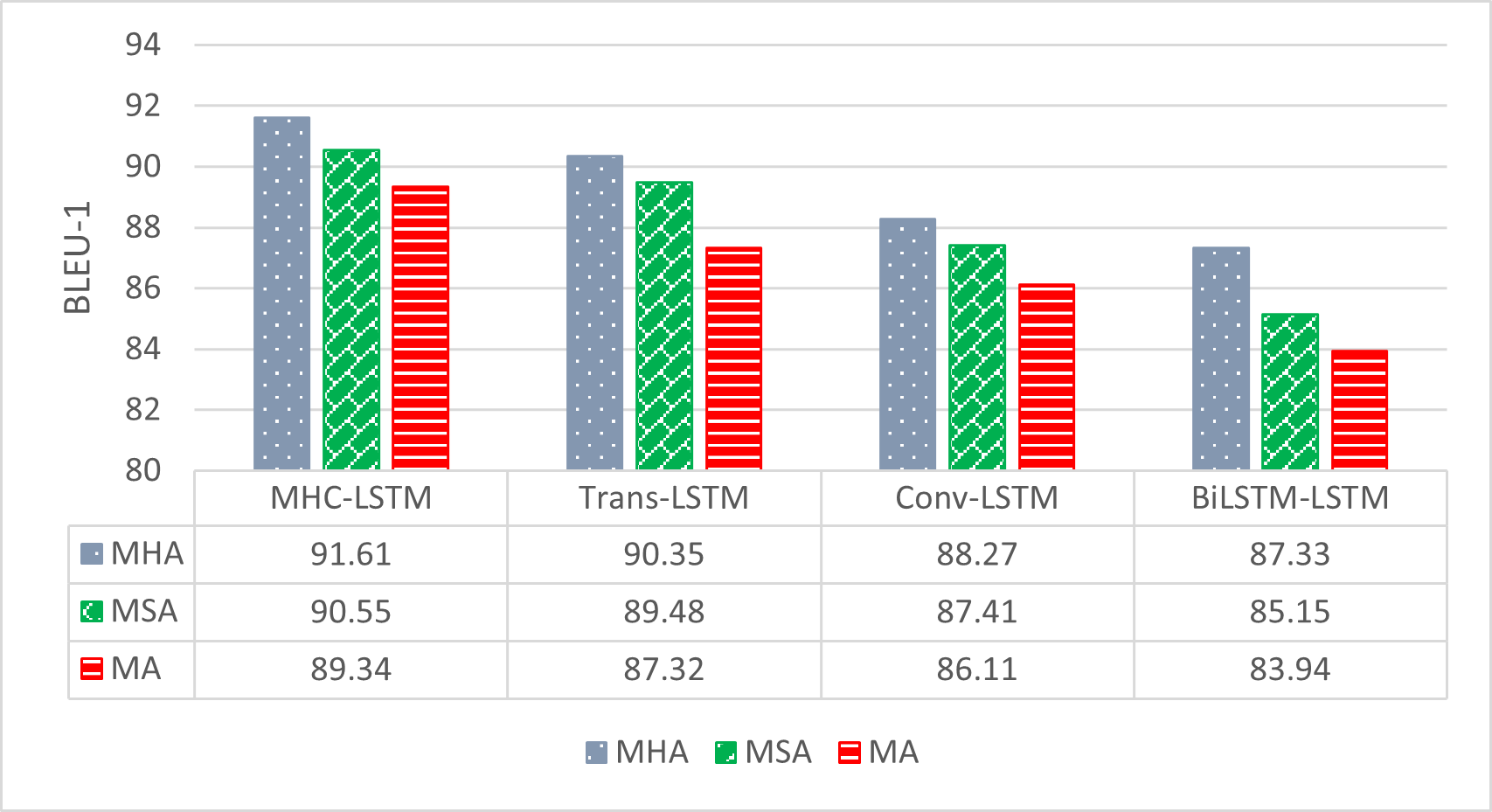}
    \caption{BLEU-1 with different cross-attention mechanisms under a fixed model architecture in LC-QuAD-1.0.}
    \label{fig4-5}
\end{figure}

\begin{figure}[H]
    \centering
    \includegraphics[width=12cm]{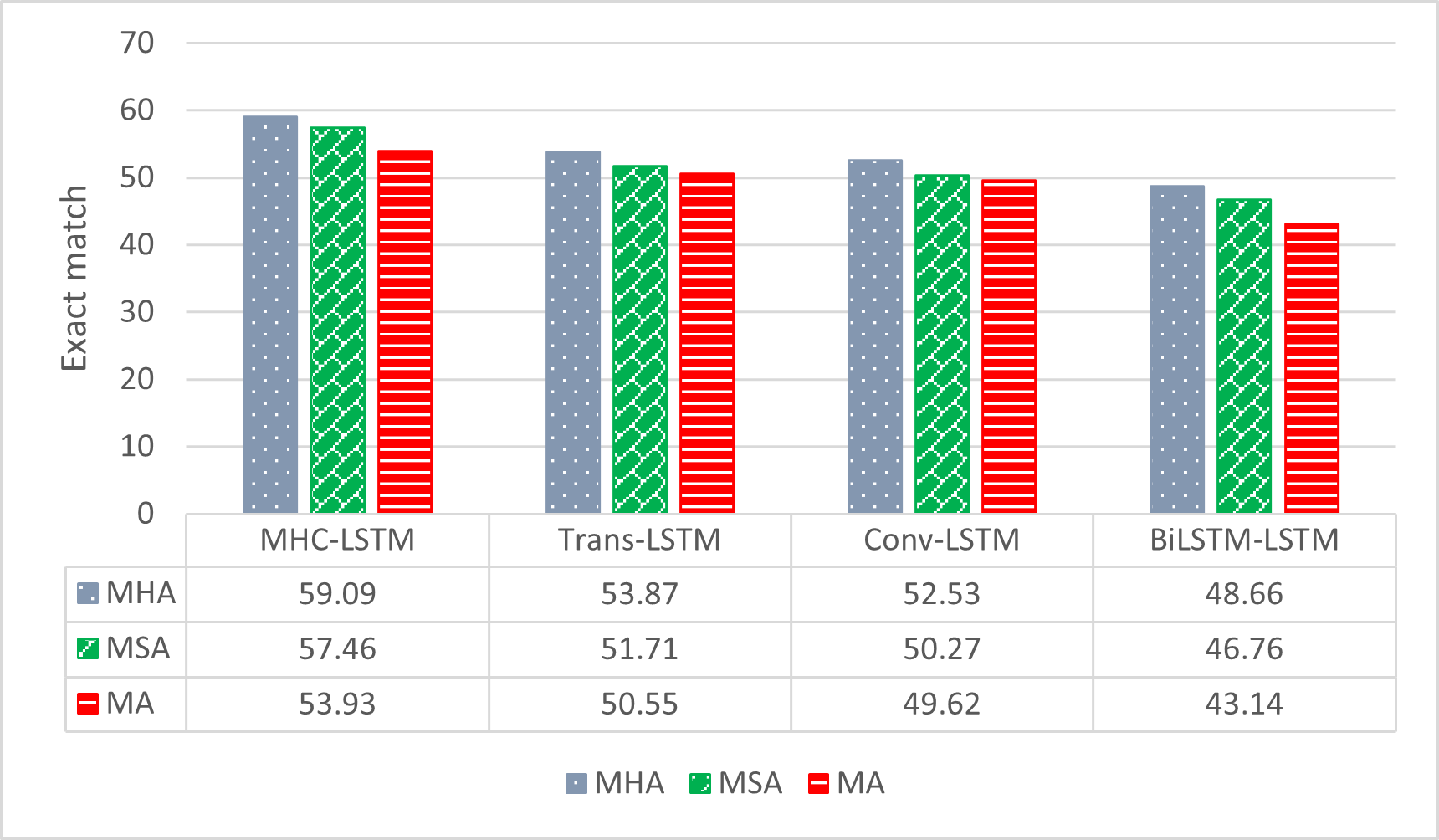}
    \caption{Exact match with different cross-attention mechanisms under a fixed model architecture in LC-QuAD-1.0.}
    \label{fig4-6}
\end{figure}

The performance results for the LC-QuAD-1.0 dataset, with a fixed model architecture and different cross-attention mechanisms, are depicted in Fig. \ref{fig4-5} and Fig. \ref{fig4-6}.

When Trans-LSTM was fixed as the model architecture, the highest BLEU-1 and Exact match scores were obtained with MHA as the cross-attention mechanism, reaching 90.35\% and 53.87\%, respectively. MHA outperformed MSA and MA by approximately 1\% in BLEU-1 and by about 2\% and 3\% in Exact match.

When Conv-LSTM was kept as the fixed model architecture, the performance with different cross-attention mechanisms closely mirrored that of Trans-LSTM. MHA as the cross-attention mechanism yielded the best results, with BLEU-1 and Exact match scores reaching 88.27\% and 52.53\%, respectively. MHA outperformed MSA and MA by approximately 1\% and 2\% in BLEU-1, and by about 2\% and 3\% in Exact match.

In the case of BiLSTM-LSTM with different cross-attention mechanisms, the best performance was consistently achieved when using MHA, with BLEU-1 and Exact match scores reaching 87.33\% and 48.66\%, respectively. MSA performed better than MA, with MA having the lowest performance.

Based on these results, it can be concluded that all three model architectures perform best when paired with MHA, followed by MSA, and perform worst when paired with MA. Among them, Trans-LSTM paired with MHA yielded the best performance. Finally, the proposed MHC-LSTM also achieved the best performance when paired with MHA in different cross-attention scenarios, reaching BLEU-1 and Exact match scores of 91.61\% and 59.09\%, respectively. The second-best was when paired with MSA, and the worst was with MA. Comparing MHC-LSTM with other hybrid architectures, it can be seen that MHC-LSTM paired with MHA achieved the best performance, which is consistent with the results from the QALD-9 experiments.


\begin{figure}[H]
    \centering
    \includegraphics[width=12cm]{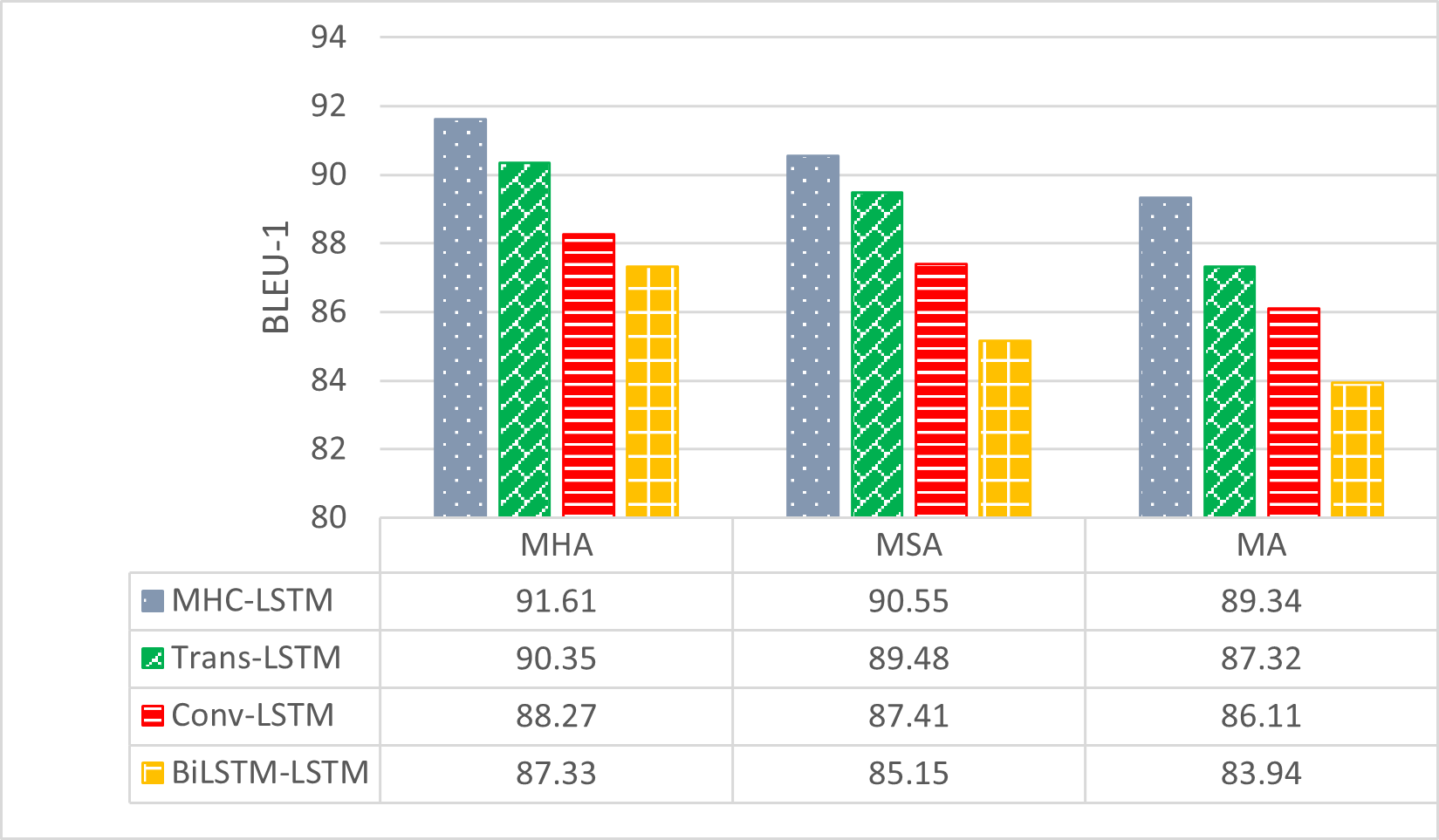}
    \caption{BLEU-1 with different model architectures under fixed cross-attention in LC-QuAD-1.0.}
    \label{fig4-7}
\end{figure}

\begin{figure}[H]
    \centering
    \includegraphics[width=12cm]{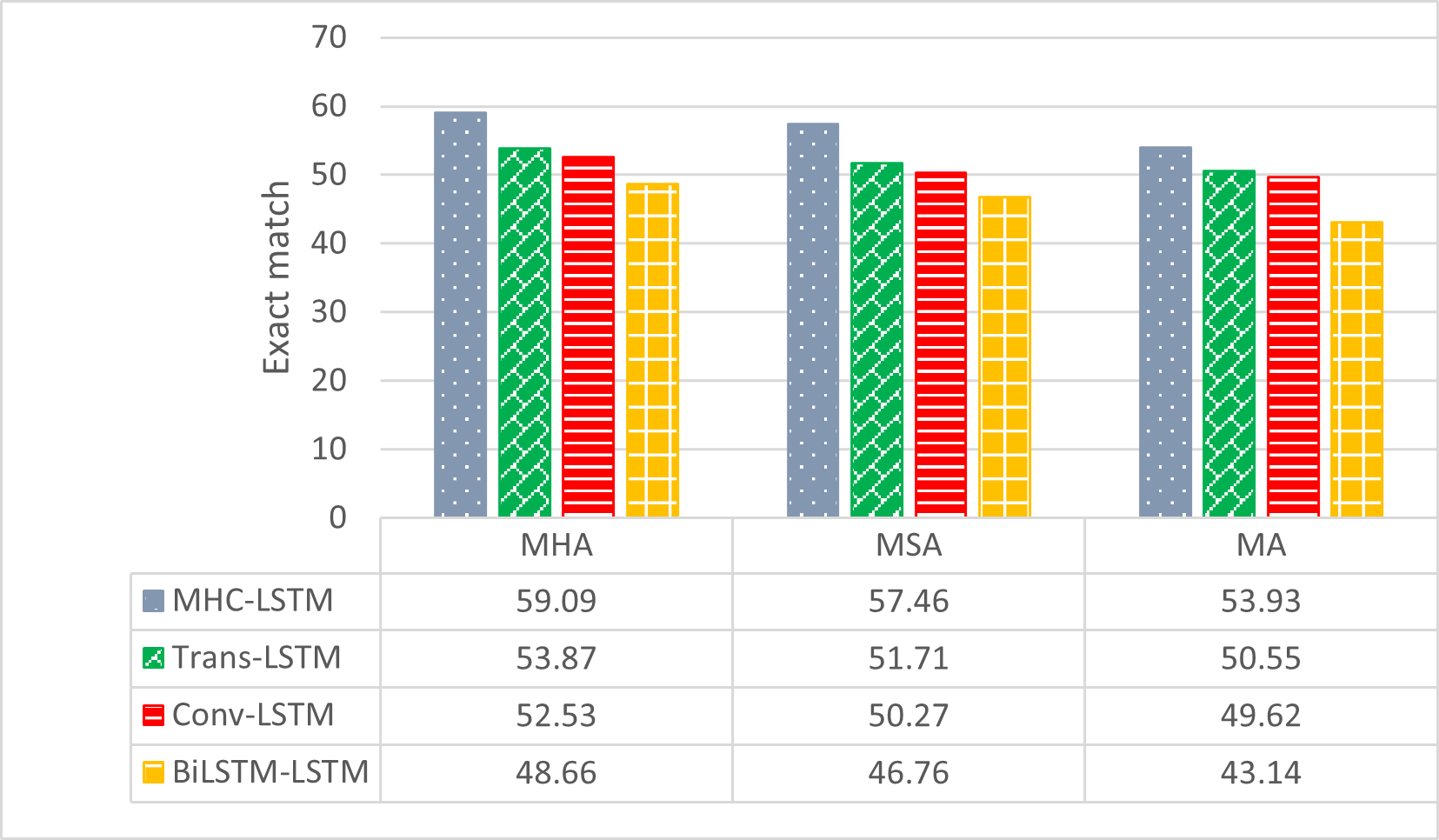}
    \caption{Exact match with different model architectures under fixed cross-attention in LC-QuAD-1.0.}
    \label{fig4-8}
\end{figure}

From another perspective, we have organized the experimental results from Fig. \ref{fig4-5} and Fig. \ref{fig4-6}, which compare the performance of different model architectures under fixed cross-attention mechanisms. The results are presented in Fig. \ref{fig4-7} and Fig. \ref{fig4-8}. 

\begin{enumerate}[(1)]
    \item MHC-LSTM consistently outperforms other models when using any Cross Attention mechanism, followed by Trans-LSTM, while Conv-LSTM and BiLSTM-LSTM exhibit comparatively lower performance.
    \item Irrespective of the encoder architecture, MHA consistently outperforms both MSA and MA, with MSA demonstrating superior performance compared to MA.
    \item The experimental results depicted in Fig. \ref{fig4-5} through Fig. \ref{fig4-8} highlight that when keeping the encoder architecture fixed, the utilization of different cross-attention mechanisms can lead to variations of up to 2 to 4\% in BLEU-1 and up to 2 to 6\% in Exact match scores. However, when maintaining a consistent cross-attention mechanism, different encoder architectures can result in more substantial differences of up to 5-6\% in BLEU-1 and as much as 10-11\% in Exact match scores. These findings underscore the significant impact of encoder architecture on overall performance.
\end{enumerate}

In summary, our findings suggest that for LC-QuAD-1.0 experiments, MHC-LSTM paired with MHA is the optimal model architecture. Regardless of the chosen architecture, MHA proves to be the superior cross-attention mechanism. These conclusions are consistent with the outcomes of the QALD-9 experiments.


In the LC-QuAD-1.0 experiments, we also examined the MHC encoder with different kernel sizes, and the outcomes are summarized in Table 4-8. Based on the results presented in Table \ref{table4-10}, employing a kernel size of 3 leads to the highest translation performance, achieving BLEU-1 and Exact match scores of 91.61\% and 59.09\%, respectively. This performance is approximately 1\% and 2\% superior to kernel sizes 5 and 7. These findings align with the results from the QALD-9 experiments, suggesting that employing a 3-gram language model can better capture the semantics of the input sequence, consistent with observations made in previous studies \cite{13, 31}.



In the performance analysis of the decoders on the LC-QuAD-1.0 dataset, we conducted a comparison between Transformer and Trans-LSTM with MHA, as well as between ConvS2S and Conv-LSTM with MSA. The summarized results can be found in Table \ref{table4-10}. When we replaced the decoder of Transformer with LSTM, there was a slight increase of 2\% in BLEU-1 and a notable 3.23\% increase in Exact match. In contrast, when we substituted the decoder of ConvS2S with LSTM, both BLEU-1 and Exact match remained unchanged. However, when MHA was employed in place of MSA with Conv-LSTM, there was an approximate 0.5\% increase in BLEU-1, and the Exact match improved by about 2\%, resulting in an overall better performance than ConvS2S. These results underscore the potential advantages of utilizing LSTM as the decoder and align with the observations made in the QALD-9 experiments. Our top-performing translation model, MHC LSTM paired with MHA, achieved the highest BLEU-1 score of 91.61\%. In comparison, Lin and Lu achieved the second-highest score of 75.06\% using a Transformer, and Yin et al. \cite{13} achieved a score of 59\% with ConvS2S. Based on the findings from the LC-QuAD-1.0 experiments, it is evident that our best translation model surpasses previous studies utilizing NMT for SPARQL translation in terms of performance.


\begin{table}[!ht]
    \centering
    \caption{Performances of Hybrid Models and Original Models in QALD-9 and LC-QuAD-1.0}
    \label{table4-10}
    \begin{tabular}{|l|l|l|l|l|}
    \hline
        Methods & Data Set & Specific Property & BLEU-1 & Exact match \\ \hline
        MHC-LSTM with MHA & QALD-9 & kernel=3 & 84.72\% & 49.17\% \\
        ~ & QALD-9 & kernel=5 & 83.18\% & 47.94\% \\ 
        ~ & QALD-9 & kernel=7 & 81.39\% & 46.28\% \\ \cline{2-5}
        & LC-QuAD-1.0 & kernel=3 & 91.61\% & 59.09\% \\ 
        & LC-QuAD-1.0 & kernel=5 & 90.78\% & 58.32\% \\
        & LC-QuAD-1.0 & kernel=7 & 89.53\% & 57.69\% \\ \hline
        Trans-LSTM with MHA & QALD-9 & None & 80.79\% & 45.28\% \\
        \cline{2-5}
         & LC-QuAD-1.0 & None & 90.25\% & 53.87\% \\ \hline
        
        Conv-LSTM with MHA & QALD-9 & None & 79.71\% & 41.95\% \\ 
        \cline{2-5}
         & LC-QuAD-1.0 & None & 88.27\% & 52.53\% \\ \hline
        
        Conv-LSTM with MSA & QALD-9 & None & 77.76\% & 40.38\% \\
        \cline{2-5}
         & LC-QuAD-1.0 & None & 87.41\% & 50.27\% \\ \hline

        ConvS2S & QALD-9 & None & 77.82\% & 40.66\% \\ 
        \cline{2-5}
        & LC-QuAD-1.0 & None & 87.72\% & 49.56\% \\ \hline
        ConvS2S (Yin et al.\cite{13}) & LC-QuAD-1.0 & None & 59\% & None \\ \hline

        Transformer & QALD-9 & None & 80.16\% & 44.13\% \\ 
         \cline{2-5}
        & LC-QuAD-1.0 & None & 88.25\% & 50.64\% \\ \hline
        Transformer(Lin and Lu \cite{15}) & LC-QuAD-1.0 & None & 75.06\% & None \\ \hline
        Transformer (Yin et al. \cite{13}) & LC-QuAD-1.0 & None & 57\% & None \\ \hline
    \end{tabular}
\end{table}

\subsection{Performances on End-to-End Question Answering System}

In our end-to-end question answering system experiments, we utilized the QALD-9 and LC-QuAD-1.0 datasets. MHC-LSTM with MHA demonstrated superior translation performance, so we employed it as the translation model within our end-to-end question answering system. To evaluate this phase, we employed metrics the Question Answering over Linked Data conference recommended, specifically Macro Recall, Macro Precision, and Macro F1-measure, summarized in Eqs. (\ref{eq4-3}), (\ref{eq4-4}), and (\ref{eq4-5}). Macro Recall quantifies the ratio of correctly identified answers to the total number of standard answers. Macro Precision, on the other hand, measures the ratio of correctly identified answers to the total number of answers retrieved by the system, including both correct and incorrect ones. The Macro F1-measure is calculated by considering both Macro Recall and Macro Precision in a relevant formula.

\begin{equation}
    \label{eq4-3}
    Marco Recall(q) = \frac{\text{number of correct system answers for }q}{\text{number of gold standard answers for }q}
\end{equation}

\begin{equation}
    \label{eq4-4}
    Marco Precision(q) = \frac{\text{number of correct system answers for }q}{\text{number of system answers for }q}
\end{equation}

\begin{equation}
    \label{eq4-5}
    Marco F1-measure = \frac{2\times Recall \ times Precision}{Recall+Precision}
\end{equation}

The performance of our research and several other end-to-end question answering systems \cite{13, 15, 16, 17, 26} on the QALD-9 dataset is summarized in Table \ref{table4-11}. Our proposed method outperforms the other systems in all three evaluation metrics. Specifically, our method achieves a Macro F1-measure of 52\%, surpassing the second-ranked system KGQAn, which scored 44\%, and the third-ranked system, Lin and Lu, with a score of 39\%. Notably, systems that utilize NMT as a foundation, including our research, KGQAn, Lin and Lu \cite{15}, and Kuo and Lu \cite{16}, consistently outperform systems using alternative techniques across all three evaluation metrics.

\begin{table}[h!]
\centering
\caption{Performance of our research and other systems on the QALD-9 end-to-end system.}
\label{table4-11}
\begin{tabular}{ |c|c|c|c| }
\hline
 Methods	&Marco Precision & Marco Recall	& Marco F1-measure
 \\ \hline
 gAnswer	& 29\%	&32\%	&29\%\\ \hline
 WDAqua-core1&26\%	&26\%	&25\%\\ \hline
QAwizard&	31\%&	47\%&	33\%\\ \hline
Lin and Lu\cite{15}&	37\%&	50\%&	39\%\\ \hline
Kuo and Lu\cite{16}&	34\%&	39\%&	36\%\\ \hline
KGQAn&	50\%&	40\%&	44\%\\ \hline
Our Proposed Approach 	& 43\% & 63\% & 52\% \\
(MHC-LSTM with MHA) &&&\\\hline
\end{tabular}
\end{table}

Next, the performances of the proposed scheme and the other end-to-end question answering systems \cite{16, 17, 23, 24} on the LC-QuAD-1.0 dataset are summarized in Table \ref{table4-12}. Similarly, the proposed method outperforms other systems in three evaluation metrics. Notably, our method achieves a Marco F1-measure of 66\%, significantly surpassing the second-ranked system KGQAn with 52\%, and the third-ranked systems DTQA and QAMP with 33\%. Furthermore, systems using NMT as a foundation, including our research and KGQAn, outperform systems using other techniques in all three evaluation metrics.

It's essential to highlight that our research attains exceptional results without relying on extensive pre-trained models like BART. This represents a substantial breakthrough, particularly in research settings constrained by limited computational resources. Furthermore, the architecture we propose in our research, MHC-LSTM with LHA, could serve as a valuable reference for developing large pre-trained models.

\begin{table}[h!]
\centering
\caption{Performance of our research and other systems on the LC-QuAD-1.0 end-to-end system.}
\label{table4-12}
\begin{tabular}{ |c|c|c|c| }
\hline
 Methods	&Marco Precision & Marco Recall	& Marco F1-measure
 \\ \hline
 DTQA	&33\%&	34\%&	33\%\\ \hline
 QAMP	&25\%&	50\%&	33\%\\ \hline
Kuo and Lu \cite{16}	&21\%	&43\%	&21\%\\ \hline
KGQAn	&58\%	&47\%&	52\%\\ \hline
Our Proposed Approach 	&63\%	&82\%	&66\%\\ 
(MHC-LSTM with MHA) &&&\\\hline
\end{tabular}
\end{table}

Given the possibility of the translation model generating inaccurate Non-Question Tokens (NQTs), we aim to comprehend the potential error types within NQTs and their associated probabilities. To achieve this, we adapted the error analysis methodology initially introduced by Banerjee et al. \cite{27}. Moreover, recognizing that incorrect question classification can result in the selection of an inappropriate Query Form and consequently lead to the retrieval of incorrect answers during the filtering stage, we also assessed the accuracy of our question classification based on question words.
Banerjee et al. proposed six errors that may occur during SPARQL translation. However, since their translation target is complete SPARQL syntax and our target is NQT, we only analyze two errors related to RDF triples: Triple Flip and Wrong Var. 

\begin{enumerate}[(1)]
    \item  Triple Flip refers to errors where the translated result has the subject and object positions reversed compared to the standard answer. For example, if the translated result is [(NER1, city, ?ans)] and the standard answer is [(?ans, city, NER1)], it is considered a Triple Flip error. 
    \item  Wrong Var indicates errors where the translation results in incorrect variables. For example, if the translated result is [(NER1, city, ?x)] and the standard answer is [(NER1, city, ?ans)], the ?x in the translation result is considered a Wrong Var error. Besides these two errors proposed by Banerjee et al., we also found errors related to the quantity of RDF triples in the translation result not matching the standard answer. We refer to this error as Wrong Quantity. For example, if the predicted result is [(NER1, related, ?ans)] and the standard answer is [(NER1, related, ?ans), (NER2, related, ?ans)], it is considered a Wrong Quantity error.
\end{enumerate}

Based on the above explanation, we performed error analysis on NQTs generated by the best translation model (MHC-LSTM) and the corrected NQTs for three error types: Triple Flip, Wrong Var, and Wrong Quantity. The results are summarized in Table \ref{table4-13}. PGN, BART, and T5-Small are the results from \cite{28}, and they did not perform analysis on Wrong Quantity. From Table \ref{table4-13}, we can observe the following results: 

\begin{enumerate}[(1)]
    \item The probability of Triple Flip in the uncorrected NQT is approximately 23\%, while in the corrected NQT, it increases to 34\%. Both error probabilities are lower than PGN and T5-Small but higher than BART. The increase in Triple Flip error probability in the corrected NQT is due to not considering the directionality of NQT during the correction process.
    \item The probability of Wrong Var in the uncorrected NQT is about 60\%, which decreases to 36\% in the corrected NQT. Both error probabilities are significantly lower than PGN's 78\% and T5-Small's 38\%, but still higher than BART's 18\%.
    \item The probability of Wrong Quantity in the uncorrected NQT is approximately 46\%, significantly reducing to 22\% in the corrected NQT. It should be noted that, while we followed the same approach as \cite{14} in randomly selecting 100 error instances, the errors selected may not be the same.
\end{enumerate}


\begin{table}[h!]
\centering
\caption{Statistical Summary of Error Types in Randomly Selected 100 NQT Translation Errors.}
\label{table4-13}
\begin{tabular}{ |c|c|c|c|c|c| }
\hline
Method&	Before NQT 	&After NQT 	&PGN	&BART	&T5-Small\\
& Modification Phase &Modification Phase&&&\\\hline
Triple Flip	&23	&34	&56	&22	&66\\\hline
Wrong Var&	60&	36&	78&	18&	36\\\hline
Wrong Quantity&	46&	24&	-&	-&	-\\\hline
\end{tabular}
\end{table}

We categorized the question types based on the interrogative words into six major categories (boolean, number, person, place, date, thing). To assess the effectiveness of this classification method, we calculated the accuracy of question classification in both QALD-9 and LC-QuAD-1.0 datasets, and the results are summarized in Table \ref{table4-14}.

Next, we will analyze the performance of this question classification method on both datasets:

\begin{enumerate}[(1)]
    \item In the QALD-9 dataset, questions belong to five question types other than boolean. The question classification achieved 100\% accuracy for number and thing questions, while it reached approximately 90\% accuracy for person, place, and date categories. Overall, our question classification method proves to be effective in the QALD-9 dataset.
    
    \item In the LC-QuAD-1.0 dataset, questions belong to five types other than date. Although our question classification method achieved 100\% accuracy for boolean and 94\% accuracy for number questions, it achieved lower accuracies of 88\%, 75\%, and 66\% for thing, person, and place categories, respectively.
\end{enumerate}


\begin{table}[h!]
\centering
\caption{The accuracy of question classification using interrogative words.}
\label{table4-14}
\begin{tabular}{ |c|c|c| }
\hline
types&	QALD-9&	LC-QuAD-1.0\\\hline
boolean&	-&	100\%\\\hline
number&	100\%	&94\%\\\hline
person&	88\%&	75\%\\\hline
place&	88\%&	66\%\\\hline
date&	91\%&	-\\\hline
thing&	100\%&	88\%\\\hline
\end{tabular}
\end{table}


\begin{table}[h!]
\centering
\caption{The performance using different separators for translation quality.}
\label{table4-15}
\begin{tabular}{ |c|c|c| }
\hline
Evaluation	& MHC-LSTM with & MHC-LSTM with \\
None & MHA (sep/sep\_end) &	MHA (comma/dot)\\ \hline
BLEU-1 &	83.37\% &	81.94\%\\ \hline
Exact match &	44.34\% &	43.56\%\\ \hline
\end{tabular}
\end{table}

Finally, to confirm whether using [sep] and [sep\_end] as the separators for NQT improves translation quality, we conducted the same experiment by following Lin and Lu's approach \cite{15} of using commas and periods as NQT separators. The results are summarized in Table \ref{table4-15}.

In the end, using [sep] and [sep\_end] as separators achieved superior performance with 83.37\% in BLEU-1, slightly outperforming the use of commas and periods, which achieved 81.94\%.

\section{Conclusions}

In this study, we analyzed the performance of hybrid model architectures (MHC-LSTM, Trans-LSTM, Conv-LSTM, BiLSTM-LSTM) in conjunction with different cross-attention mechanisms on the task of transforming natural language questions into SPARQL queries using the QALD-9 dataset \cite{14} and LC-QuAD-1.0 dataset. We found that all hybrid encoder-decoder structures performed best when combined with MHA \cite{14}, followed by MSA \cite{19}, while MA \cite{18} yielded inferior results. MHC-LSTM achieved the best results as a fixed architecture, achieving excellent BLEU-1 of 83.37\% and an Exact match of 44.34\% on the LC-QuAD-1.0 dataset. The research results also indicate that our proposed optimal translation model outperformed Transformer and ConvS2S, suggesting that hybrid model architectures with different cross-attention mechanisms are worth exploring.

Additionally, this study proposed using NQT as the output of the translation model and introduced an NQT inspection and correction mechanism to modify the translated NQTs. The research showed that the NQT inspection and correction mechanism significantly improved translation quality. For example, with the optimal translation model, BLEU-1 and Exact match improved from 83.37\% and 44.34\% to 91.61\% and 59.09\%, respectively.

While this study made significant progress in SPARQL translation tasks, there are still areas worth exploring. First, the experimental results indicate that the architecture of the encoder, decoder, and the calculation method of cross-attention all impact translation task effectiveness. Although our proposed MHC-LSTM shows substantial improvements, different encoder-decoder structures are still worth investigating. Furthermore, the NQT correction mechanism proposed in this study deserves further exploration. Since this mechanism heavily relies on the accuracy of the Multiple Entity Type Tagger (METT), future work should focus on improving the performance of METT or exploring alternative solutions.

\bibliographystyle{apalike-refs}
\bibliography{sample}

\end{document}